\documentclass{article}

\usepackage[final,nonatbib]{neurips_2021}

\usepackage[utf8]{inputenc} % allow utf-8 input
\usepackage[T1]{fontenc}    % use 8-bit T1 fonts
\usepackage{hyperref}       % hyperlinks
\usepackage{url}            % simple URL typesetting
\usepackage{booktabs}       % professional-quality tables
\usepackage{amsfonts}       % blackboard math symbols
\usepackage{nicefrac}       % compact symbols for 1/2, etc.
\usepackage{microtype}      % microtypography
\usepackage{xcolor}         % colors

\usepackage{microtype}
\usepackage{graphicx}
\usepackage{subfigure}
\usepackage{amsmath,amssymb}
\usepackage[normalem]{ulem}
\usepackage{xstring}
\usepackage{mathtools, nccmath}
\usepackage{xpatch}
\usepackage{amsthm}
\usepackage{siunitx}
\usepackage{hyperref}
\usepackage{wrapfig}

\newcommand{\benchmark}{Continual World}
\newcommand{\metaworld}{Meta-World}

\newcommand{\piotrmn}[1]{\textcolor{blue}{\small [pm(nice to have): #1]}}
\renewcommand{\piotrmn}[1]{}

\definecolor{darkgreen}{rgb}{0., 0.5, 0.}

\newcommand{\cwtwenty}{CW20}

\newtheorem{theorem}{Theorem}
\newtheorem{fact}[theorem]{Fact}

\title{Continual World: A Robotic Benchmark For Continual Reinforcement Learning}

\author{%
  Maciej Wołczyk\thanks{equal contribution} \\
  Jagiellonian University\\
  Kraków, Poland \\
  \texttt{maciej.wolczyk@doctoral.uj.edu.pl} \\
  \And
  Michał Zając$^*$ \\ % if anybody knows how to do this properly, please change.
  Jagiellonian University\\
  Kraków, Poland \\
  \texttt{emzajac@gmail.com} \\
  \AND
  Razvan Pascanu \\
  DeepMind \\
  London, UK \\
  \texttt{razp@google.com} \\
  \And
  Łukasz Kuciński \\
  Polish Academy of Sciences \\
  Warsaw, Poland \\
  \texttt{lkucinski@impan.pl} \\
  \And
  Piotr Miłoś \\
  Polish Academy of Sciences, \\
  University of Oxford,\\
  deepsense.ai\\
  Warsaw, Poland \\
  \texttt{pmilos@impan.pl} \\
}

\begin{document}

\maketitle

\begin{abstract}
  Continual learning (CL) --- the ability to continuously learn, building on previously acquired knowledge --- is a natural requirement for long-lived autonomous reinforcement learning (RL) agents. While building such agents, one needs to balance opposing desiderata, such as constraints on capacity and compute, the ability to not catastrophically forget, and to exhibit positive transfer on new tasks. Understanding the right trade-off is conceptually and computationally challenging, which we argue has led the community to overly focus on \emph{catastrophic forgetting}.  In response to these issues, we advocate for the need to prioritize forward transfer and propose \emph{Continual World}, a benchmark consisting of realistic and meaningfully diverse robotic tasks built on top of Meta-World \cite{metaworld} as a testbed. Following an in-depth empirical evaluation of existing CL methods, we pinpoint their limitations and highlight unique algorithmic challenges in the RL setting. Our benchmark aims to provide a meaningful and computationally inexpensive challenge for the community and thus help better understand the performance of existing and future solutions. Information about the benchmark, including the open-source code, is available at \url{https://sites.google.com/view/continualworld}.
\end{abstract}

% \piotrm{Camera ready to do list:}
% \begin{itemize}
%     \item time and memory table in App B.4
%     \item more realistic robotics tasks Rev EkWJ (ignore?)
%     \item Concerning PackNet (in triplet experiments), we note that even though the parameters are isolated/frozen, there is interaction via the activations (activations related to the previous tasks are not masked and can be reused). We will provide a more detailed description in the camera-ready version.
%     \item Give a final read
%     \item Review webpage include info about being accepted to neurips
%     \item Put the new version on arXiv 
%     \item update github with information about being accepted to NeurIPS 
%     \item submit code to NeurIPS ? (rather not)
% \end{itemize}

\section{Introduction}\label{sec:introduction}
Change is ubiquitous. Unsurprisingly, due to evolutionary pressure, humans can quickly adapt and creatively reuse their previous experiences. In contrast, although biologically inspired, deep learning (DL) models excel mostly in static domains that satisfy the i.i.d. assumption, as for example in image processing~\cite{DBLP:conf/nips/KrizhevskySH12,efficientnet,biggan,dalle}, language modelling~\cite{DBLP:conf/nips/VaswaniSPUJGKP17, DBLP:conf/nips/BrownMRSKDNSSAA20} or biological applications  \cite{DBLP:journals/nature/Senior0JKSGQZNB20}. As the systems are scaled up and deployed in open-ended settings, such assumptions are increasingly questionable; imagine, for example, a robot that needs to adapt to the changing environment and the wear-and-tear of its hardware. \textit{Continual learning} (CL), an area that explicitly focuses on such problems, has been gaining more attention recently. The progress in this area could offer enormous advantages for deep neural networks \cite{HADSELL20201028} and move the community closer to the long-term goal of building intelligent machines \cite{HASSABIS2017245}.

Evaluation of CL methods is challenging. Due to the sequential nature of the problem that disallows parallel computation, evaluation tends to be expensive, which has biased the community to focus on toy tasks. These are mostly in the domain of supervised learning, often relying on MNIST. In this work, we expand on previous discussions on the topic \cite{Schwarz2018, DBLP:journals/corr/abs-1805-09733, DBLP:conf/cvpr/0001DCM20, DBLP:conf/icml/Schwarz0LGTPH18}  and introduce a new benchmark, \textit{Continual World}. The benchmark is built on realistic robotic manipulation tasks from Meta-World \cite{metaworld}, benefiting from its diversity but also being computationally cheap. Moreover, we provide shorter auxiliary sequences, all of which enable a quick research cycle. On the conceptual level, a fundamental difficulty of evaluating CL algorithms comes from the different desiderata for a CL solution. These objectives are often opposing each other, forcing practitioners to explicitly or implicitly make trade-offs in their algorithmic design that are data-dependent. \textit{Continual World} provides more meaningful relationships between tasks, answering recent calls \cite{HADSELL20201028} to increase attention on forward transfer.

Additionally, we provide an extensive evaluation of a spectrum of commonly used CL methods. It highlights that many approaches can deal relatively well with \textit{catastrophic forgetting} at the expense of other desiderata, in particular forward transfer. This emphasizes our call for focusing on \textit{forward transfer} and the need for more benchmarks that allow for common structure among the tasks.

The main contribution of this work is a CL benchmark that poses optimizing forward transfer as the central goal and shows that existing methods struggle to outperform simple baselines in terms of the forward transfer capability. We release the code\footnote{\url{https://github.com/awarelab/continual_world}} both for the benchmark and $7$ CL methods, which aims to provide the community helpful tools to better understand the performance of existing and future solutions. We encourage to visit the website\footnote{\url{https://sites.google.com/view/continualworld/home}} of the project and participate in the Continual World Challenge.

\section{Related work}
The field of continual learning has grown considerably in the past years, with numerous works forming new subfields \cite{DBLP:conf/nips/JavedW19} and finding novel applications \cite{DBLP:conf/iclr/SunHL20}. For brevity, we focus only on the papers proposing RL-based benchmarks and point to selected surveys of the entire field. \cite{HADSELL20201028} provide a high-level overview of CL and argue that learning in a non-stationary setting is a fundamental problem for the development of AI, highlighting the frequent connections to neuroscience. On the other hand, \cite{de2019continual, DBLP:journals/nn/ParisiKPKW19} focus on describing, evaluating, and relating CL methods to each other, providing a taxonomy of CL solutions that we use in this work. 

The possibility of applying CL methods in reinforcement learning scenarios has been explored for a long time, see \cite{khetarpal2020continual} for a recent review. However, no benchmark has been widely accepted by the community so far, which is the aim of this work. Below we discuss various benchmarks and environments considered in the literature.

\textbf{Supervised settings} MNIST has been widely used to benchmark CL algorithms in two forms~\cite{DBLP:journals/corr/KirkpatrickPRVD16}. In the permuted MNIST, the pixels of images are randomly permuted to form new tasks. In the split MNIST, tasks are defined by classifying non-overlapping subsets of classes, e.g. $0$ vs. $1$ followed by $2$ vs. $3$. A similar procedure has been applied to various image classification tasks like CIFAR-10, CIFAR-100, Omniglot or mini-ImageNet~\cite{DBLP:conf/iclr/Adel0T20,DBLP:conf/icml/Schwarz0LGTPH18,DBLP:conf/nips/AljundiBTCCLP19}. Another benchmark is CORe50~\cite{DBLP:conf/corl/LomonacoM17}, a dataset for continuous object recognition. Recent work~\cite{kruszewski2021evaluating} proposes a benchmark based on language modeling. We find that many of these benchmarks are challenging and allow to measure forgetting. However, we argue they are not geared towards measuring forward transfer or for highlighting important RL-specific characteristics of the CL problem.

\textbf{Atari} The Atari 2600 suite~\cite{bellemare13arcade} is a widely accepted RL benchmark. Sequences of different Atari games have been used for evaluating continual learning approaches~\cite{DBLP:journals/corr/RusuRDSKKPH16,DBLP:journals/corr/KirkpatrickPRVD16}. Using Atari can be computationally expensive, e.g., training a sequence of ten games typically requires $100$M steps or more. More importantly, as \cite{DBLP:journals/corr/RusuRDSKKPH16} notes, these games lack a meaningful overlap, limiting their relevance for studying transfers. \textbf{Continuous control} \cite{DBLP:journals/corr/abs-1912-01188,pmlr-v97-kaplanis19a} use continuous control tasks such as Humanoid or Walker2D. However, the considered sequences are short, and the range of experiments is limited. \cite{DBLP:conf/nips/MendezWE20} use \metaworld{} tasks, similarly to us, for evaluations of their continual learning method, but the work is not aimed at building a benchmark. As such, it uses the \metaworld{}'s MT10 preset and does not provide an in-depth analysis of the tasks or other CL methods. \textbf{Maze navigation} A set of 3D maze environments is used in \cite{DBLP:journals/corr/RusuRDSKKPH16}. The map structure and objects that the agent needs to collect change between tasks. It is not clear, though, if the tasks provide enough diversity. \cite{DBLP:conf/cvpr/0001DCM20} propose CRLMaze, 3D navigation scenarios for continual learning, which solely concentrate on changes of the visual aspects. \textbf{StarCraft} \cite{Schwarz2018} present a StarCraft campaign ($11$ tasks) to evaluate a high-level transfer of skills. The main drawback of this benchmark is excessive computational demand (often more than $1$B frames). \textbf{Minecraft} \cite{DBLP:conf/aaai/TesslerGZMM17} propose simple scenarios within the Minecraft domain along with a hierarchical learning method. The authors phrase the problem as lifelong learning and do not use typical CL methods.
\textbf{Lifelong Hanabi} \cite{DBLP:conf/icml/NekoeiBCC21} consider a multi-agent reinforcement learning setting based on Hanabi, a cooperative game requiring significant coordination between agents. On the other hand, we focus on the single agent setting with changing environment, which allows us to bypass the computational complexity needed to model interactions between agents and highlight issues connected to learning in a changing world. \textbf{Causal World} \cite{DBLP:conf/iclr/AhmedTGNWBSB21} propose an environment for robotic manipulation tasks which share causal structure. Although they investigate issues deeply connected to learning in a changing world, such as generalization to new tasks and curricula, they do not directly consider continual learning.
\textbf{Jelly Bean World} \cite{DBLP:conf/iclr/PlataniosSM20} provide interesting procedurally generated grid world environments. The suite is configurable and can host a non-stationary setting. It is unclear, however, if such environments reflect the characteristics of real-world challenges.

\section{Continual learning background}\label{sec:continual_learning_definitions}
% \subsection{Continual learning} 
Continual learning (CL) is an area of research which focuses on building algorithms capable of handling non-stationarity. They should be able to sequentially acquire new skills and solve novel tasks without forgetting the previous ones. Such systems are desired to accommodate over extended periods swiftly, which is often compared to human capabilities and alternatively dubbed as lifelong learning. CL is intimately related to multi-task learning, curriculum learning, meta-learning, with some key differences. Multi-task assumes constant access to all tasks, thus ignoring non-stationarity. Curriculum learning focuses on controlling the task ordering and often the learning time-span. Meta-learning, a large field of its own, sets the objective to develop procedures that allow fast adaptation within a task distribution and usually ignores the issue of non-stationarity.

The CL objective is operationalized by the training and evaluation protocols. The former typically consists of a sequence of tasks (their boundaries might be implicit and smooth). The latter usually involves measuring \textit{catastrophic forgetting}, \textit{forward transfer}, and \textit{backward transfer}. The learning system might also have constrained resources: \textit{computations}, \textit{memory}, \textit{size of neural networks}, and the \textit{volume of data samples}. A fundamental observation is that the above aspects and desiderata are conflicting. For example, given unlimited resources, one might mitigate forgetting simply by storing everything in memory and paying a high computational cost of rehearsing all samples from the past.
% \begin{figure}[ht!]
  \begin{wrapfigure}{l}{0.5\linewidth}
    % \vskip -0.4in
    \begin{center}
        % \centerline{\includegraphics[width=.7\columnwidth]{imgs/positive_transfer_graph_example.png}}
        \centerline{
        \includegraphics[width=.28\columnwidth]{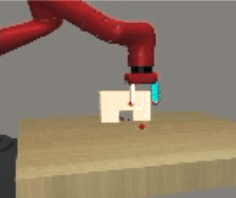}
        \hspace{-0.1in}
        \includegraphics[width=.245\columnwidth]{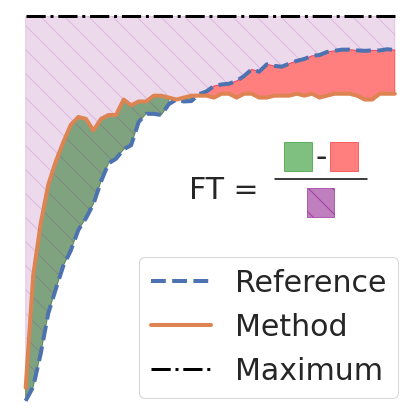}
        }
        \caption{\small Left graph shows task \textsc{peg-unplug-side-v1} and the right graph presents forward transfer from \textsc{shelf-place-v1} to \textsc{peg-unplug-side-v1}. In this case $FT=0.10$. }
        \label{fig:positive_transfer_graph}
    \end{center}
    \vskip -0.4in
  % \end{figure}
  \end{wrapfigure}

Another pair of objectives that are problematic for current methods are forgetting and forward transfer. For neural networks, existing methods propose to limit network plasticity. These alleviate the problem of forgetting, however, at the cost of choking the further learning process. We advocate for more nuanced approaches. Importantly, to make the transfer possible, our benchmark is composed of related tasks. We also put modest bounds on resources. This requirement is in line with realistic scenarios, demanding computationally efficient adaptation and inference. In a broader sense, we hope to address a data efficiency challenge, one of the most significant limitations of the current deep (reinforcement) learning methods. We conjecture that forward transfer might greatly improve the situation and possibly one day enable us to create systems with human-level cognition capabilities, in line with similar thoughts expressed in~\cite{HADSELL20201028}.

\section{\benchmark{} benchmark}
\benchmark{} is a new benchmark designed to be a testbed for evaluating RL agents on the challenges advocated by the CL paradigm,  described in Section \ref{sec:continual_learning_definitions}, as well as highlighting the RL-specific algorithmic challenges for CL (see Section \ref{sec:rl_algorithmic_challenges}). As such it is aimed at being valuable to both the CL and RL communities. \benchmark{} consists of realistic robotic manipulation tasks, aligned in a sequence to enable the study of forward transfer. It is designed to be challenging while computationally accessible.\footnote{We use $8$-core machines without GPU. Training the CW20 sequence of twenty tasks takes about $100$ hours. We also provide shorter $10$ and $3$ task sequences to speed up the experimental loop further.} The benchmark is based on \metaworld{}, a suite of robotic tasks already established in the community. This enables easy comparisons with the related fields of multi-task and meta-learning reinforcement learning, potentially highlighting one benefit of CL framing, namely that of dealing with different reward scales as we discuss more in detail in Appendix \ref{sec:mult_task_learning_appendix}. \benchmark{} comes with open-source code that allows for easy development and testing of new algorithms and provides  implementations of $7$ existing algorithms. Finally, it allows highlighting RL-specific challenges for the CL setting. We believe that our work is a step in the right direction towards reliable benchmarks of CL. We realize, however, that it will need to evolve as the field progresses. We leave a discussion on future directions and limitations to Section~\ref{sec:future_steps}.

\subsection{Metrics} \label{sec:metrics}
To facilitate further discussion, we start with defining metrics. These are rather standard in the CL setting \cite{DBLP:journals/corr/abs-1810-13166}. Assume $p_i(t)\in[0,1]$ to be the performance (success rate) of task $i$ at time $t$. As a measure of performance, we take the average success rate of achieving a goal specified by a given task when using randomized initial conditions and stochastic policies (see also Section \ref{sec:training_details}).\footnote{Stochastic evaluations are slightly more smooth and have little difference to the deterministic ones.}
Each task is trained for $\Delta=1M$ steps. The main sequence has $N=20$ tasks and the total sample budget is $T=N\cdot \Delta = 20M$. The $i$-th task is trained during the interval $t\in [(i-1) \cdot \Delta, i\cdot \Delta]$. We report the following metrics:

\textbf{Average performance}. The average performance at time $t$ is (see Figure \ref{fig:training_graph}) \useshortskip
\begin{equation}
    \text{P}(t) := \frac{1}{N} \sum_{i=1}^{N} p_i(t).\label{eq:performance}
\end{equation}
Its final value, $\text{P}(T)$, is a traditional metric used in the CL research. This is the objective we use for tuning hyperparameters. We have $\text{P}(t) \in [0, 1]$ for each $t$.

\textbf{Forward transfer}.
We measure the forward transfer of a method as the normalized area between its training curve and the training curve of the reference, single-task, experiment, see Figure \ref{fig:positive_transfer_graph}. Let $p_i^b\in[0,1]$ be the reference performance\footnote{Note that we avoid trivial tasks, for which $\text{AUC}_i^b=1$ is making the metric ill-defined. Additionally, we acknowledge the dependency on the hyperparameters of the learning algorithm and that there are alternative quantities of interest like relative improvement in performance rather than faster learning.} then the forward transfer for the task $i$, denoted by $\text{FT}_i$, is \useshortskip
\begin{align*}
    \text{FT}_i    & := \frac{\text{AUC}_i-\text{AUC}_i^b}{1-\text{AUC}_i^b},\quad                        
    \text{AUC}_i   := \frac{1}{\Delta}\int_{(i-1)\cdot\Delta}^{i \cdot \Delta} p_i(t) \text{d}t, \quad 
    \text{AUC}_i^b  :=  \frac{1}{\Delta}\int_{0}^{\Delta} p_i^b(t) \text{d}t,
\end{align*}
The average forward transfer for all tasks, $\text{FT}$, is defined as\useshortskip
\begin{equation}\label{eq:forward_transfer}
    \text{FT} = \frac{1}{N}\sum_{i=1}^N \text{FT}_i.
\end{equation}
We note that $\text{FT}_i \leq 1$ and they might be negative. In our experiments, we also measure backward transfer. As it is negligible, see Appendix \ref{sec:forgetting_appendix}.

\textbf{Forgetting}. For task $i$, we measure the decrease of performance after ending its training, i.e.\useshortskip
\begin{equation}
    F_i = p_i(i \cdot \Delta) - p_i(T). \label{eq:forgetting}
\end{equation}
Similarly to $\text{FT}$, we report  $F = \frac{1}{N}\sum_{i=1}^N F_i$. We have $F_i\leq 1$ for any $i$ and consequently $\text{FT}\leq 1$. It is possible that $F_i$ are negative, which would indicate backward transfer. We do not observe this in practice, see Appendix \ref{sec:forgetting_appendix}.

\subsection{Continual World tasks}\label{sec:task_sets}
This section describes the composition of Continual World benchmark and the rationale behind its design. We decided to base on \metaworld{} \cite{metaworld}, a fairly new but already established robotic benchmark for multi-task and meta reinforcement learning. From a practical standpoint, \metaworld{} utilizes the open-source MuJoCo physics engine \cite{DBLP:conf/iros/TodorovET12}, prized for speed and accuracy. \metaworld{} provides $50$ distinct manipulation tasks with everyday objects using a simulated robotic Sawyer arm. Although the tasks vary significantly, the structure and semantics of observation and action spaces remain the same, allowing for transfer between tasks. Each observation is a $12$-dimensional vector containing $(x, y, z)$ coordinates of the robot's gripper and objects of interest in the scene. The $4$-dimensional action space describes the direction of the arm's movement in the next step and the gripper actuator delta. Reward functions are shaped to make each task solvable. In evaluations, we use a binary \textit{success metric} based on the distance of the task-relevant object to its goal position. This metric is interpretable and enables comparisons between tasks. For more details about the rewards and evaluation metrics, see \cite[Section 4.2, Section 4.3]{metaworld}.

\textbf{\cwtwenty{}, CW10, triplets sequences}
The core of our benchmark is \cwtwenty{} sequence. Out of 50 tasks defined in \metaworld{}, we picked those that are not too easy or too hard in the assumed sample budget $\Delta = 1M$. Aiming to strike a balance between the difficulty of the benchmark and computational requirements, we selected $10$ tasks. The tasks and their ordering were based on the transfer matrix (see the next paragraph), so that there is a high variation of forward transfers (both in the whole list and locally).
We refer to these ordered tasks as CW10, and \cwtwenty{} is CW10 repeated twice. We recommend using \cwtwenty{} for final evaluation; however,  CW10 is already very informative in most cases. Due to brevity constraints, we present an ablation with an alternative ordering of the tasks and a longer sequence of $30$ tasks in Appendix \ref{sec:ablations_appendix}, however, these experiments do not alter our findings. Additionally, to facilitate a fast development cycle, we propose a set of triplets, sequences of three tasks which exhibit interesting learning dynamics.

The CW10 sequence is: {\small \textsc{hammer-v1}, \textsc{push-wall-v1}, \textsc{faucet-close-v1}, \textsc{push-back-v1}, \textsc{stick-pull-v1},
\textsc{handle-press-side-v1}, \textsc{push-v1}, \textsc{shelf-place-v1}, \textsc{window-close-v1}, \textsc{peg-unplug-side-v1}}.

\begin{wrapfigure}{l}{0.5\linewidth}
    % \vskip -0.4in
    \begin{center}
        %\centerline{\includegraphics[width=\columnwidth]{imgs/transfer_matrix_sorted.png}}
        % \centerline{\includegraphics[width=.85\columnwidth]{imgs/heatmap_rg.png}}
        % \centerline{\includegraphics[width=.9\columnwidth]{imgs/graph_transfer_v2.png}}
        \centerline{
        \includegraphics[width=.45\columnwidth]{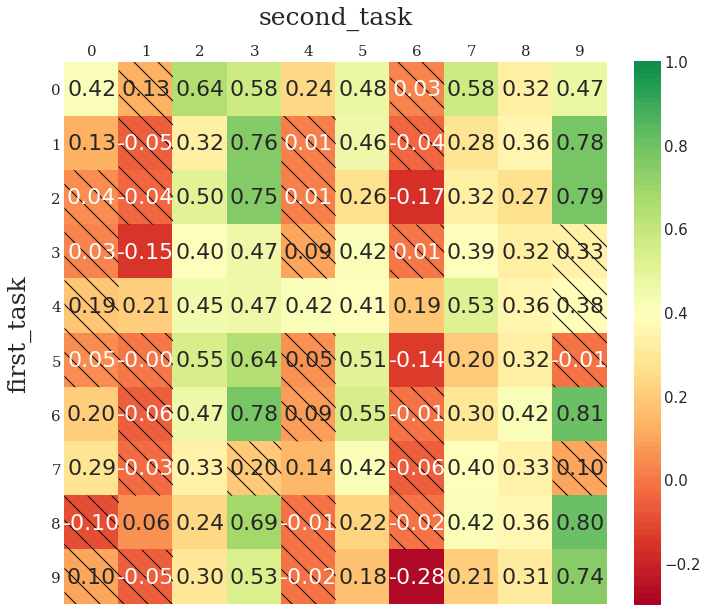}
        \hskip 0.2in
        % \raisebox{0.7\height}{\includegraphics[width=.45\columnwidth]{imgs/graph_transfer_v2.png}}
        }
        \caption{\small
            Transfer matrix, see Section \ref{sec:task_sets}. Each cell represents the forward transfer from the first task to the second one. We shaded the cells for which $0$ belongs to their $90\%$ confidence interval. %Right: graph visualization.
        }\label{fig:transfer_matrix_haatmmap}
    \end{center}
    \vskip -0.5in
  \end{wrapfigure}

\textbf{Transfer matrix} Generally, the relationship between tasks and its impact on learning dynamics of neural networks is hard to quantify, where semantic similarity does not typically lead to transfer~\cite{yunshu}. To this end, we consider a minimal setting, in which we finetune on task $t_2$ a model pretrained on $t_1$, using the same protocol as the benchmark (e.g., different output heads, see Section~\ref{sec:training_details}). This provides neural network-centric insight into the relationship between tasks summarized in Figure \ref{fig:transfer_matrix_haatmmap}, and allows us to measure \emph{low-level transfer} between tasks, i.e., the ability of the model to reuse previously acquired features. See  Appendix \ref{sec:forward_transfer_matrices_appendix} for more results and extended discussion.

% \begin{figure}[h!]
% \end{figure}

Notice that there are only a few negative forward transfer cases, and those are of a rather small magnitude (perhaps unsurprisingly, as the tasks are related). There are also visible patterns in the matrix. For instance, some tasks such as \textsc{peg-unplug-side-v1} or \textsc{push-back-v1} benefit from a relatively large forward transfer, (almost) irrespective of the first task. Furthermore, the average forward transfer given the second task (columns) is more variable than the corresponding quantity for the first task (rows). 

Note that some transfers on the diagonal (i.e., between the same tasks) are relatively small. We made a detailed analysis of possible reasons, which revealed that the biggest negative impact is due to the replay buffer resets, which seems, however, unavoidable for off-diagonal cases, see Section \ref{sec:rl_algorithmic_challenges} for details.

Importantly, we use this matrix to estimate what level of forward transfer a good CL method should be able to achieve. We expect that a model which is able to remember all meaningful aspects of previously seen tasks would transfer at least as well as if one were just fine-tuning after learning the best choice between the previous tasks. For a sequence $t_1, \ldots, t_N$ we set the reference forward transfer, $\text{RT},$ to be
\begin{equation}
    \text{RT}:=\frac{1}{N}\sum_{i=2}^N \max_{j<i} \text{FT}(t_j, t_i) \label{eq:forward_transfer_upperbound},
\end{equation}
where $\text{FT}(t_j, t_i)$ is the transfer matrix value for $t_j, t_i$. For the \cwtwenty{} sequence, the value is $\text{RT}=0.46$. Note that a \emph{model can do better} that this by composing knowledge from multiple previous tasks.

\subsection{Training and evaluation details} \label{sec:training_details}

We adapt the standard \metaworld{} setting to CL needs. First, we use separate policy heads for each task, instead of the original one-hot task ID inputs (we provide ablation experiments for this choice in Appendix \ref{sec:ablations_appendix}). Second, in each episode, we randomize the positions of objects in the scene to encourage learning more robust policies. We use an MLP network with $4$ layers of $256$ neurons.

For training, we use soft actor-critic (SAC)  \cite{DBLP:conf/icml/HaarnojaZAL18}, a popular and efficient RL method for continuous domains.  SAC is an off-policy algorithm using replay buffer, which is an important aspect for CL, particularly for methods relying on rehearsing old trajectories. SAC is based on the so-called maximum entropy principle; this results in policies which explore better and are more robust to changes in the environment dynamics. Both of these qualities might be beneficial in CL.% to facilitate task transitions.

We note that the size of the neural network and optimization details of the SAC algorithm (like batch size) put constraints on "the amount of compute". Intentionally, these are rather modest, which is in line with CL desiderata, see Section~\ref{sec:continual_learning_definitions}. Similarly, we limit the number of timesteps to $1M$, which is a humble amount for modern-day deep reinforcement learning. We picked tasks to be challenging but not impossible within this budget. We note that training in the RL setting tends to be less stable than in the supervised one. We recommend using multiple seeds, in our experiments, we typically used $20$ and calculate confidence intervals; we used the bootstrap method. We choose hyperparameters that maximize average performance \eqref{eq:performance}. In our experiments, we tune common parameters for SAC and the method-specific hyperparameters separately.
All details of the training and evaluation setup are presented in  Appendix \ref{sec:technical_details}.

\begin{figure*}[t]
    %\vskip 0.2in
    \begin{center}
        % \centerline{\includegraphics[width=\textwidth]{imgs/main_plot_v1_2mtl.png}}
        \centerline{\includegraphics[width=1\textwidth]{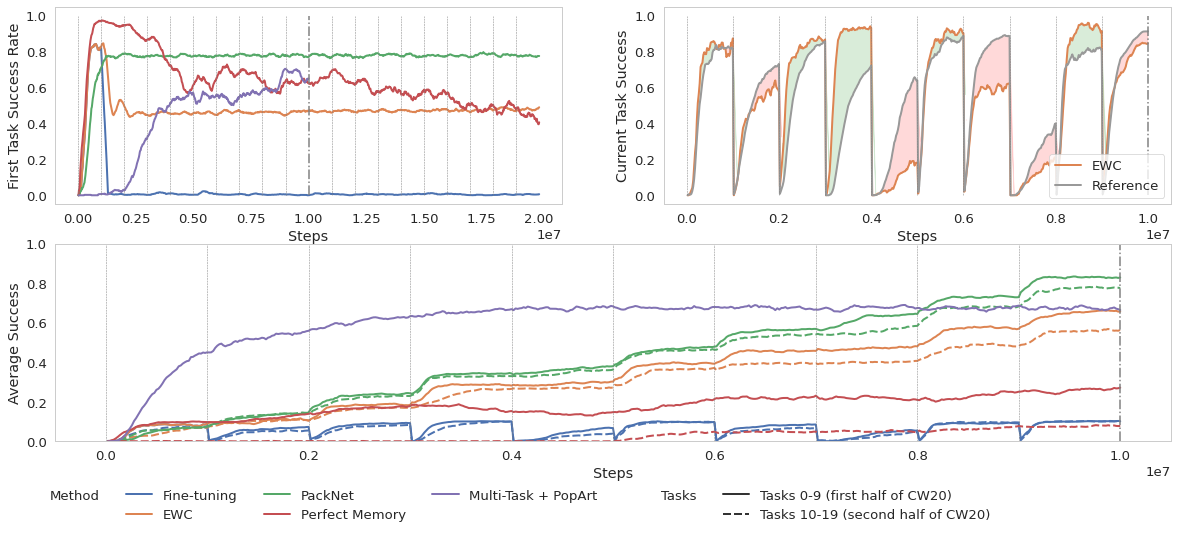}}
        \caption{{\small Training curves for selected CL methods and multi-task. The upper left panel shows the performance on the first task for a subset of methods throughout the whole training. Note that due to the use of different output heads, we do not see a second bump when revisiting this task at time $10M$. The upper right panel shows the performance on the current task being trained for EWC compared to a reference (a model learning only that task from scratch). The bottom plot shows the average performance. Solid lines show the performance of the model training on the first $10$ tasks (where $1$ means being able to solve all of them). Dashed lines show the performance of learning the same tasks in the second half of the benchmark. Note that dashed lower are below solid ones, indicating lower performance on the second pass, even if the agent has already previously learned the tasks and has access to relevant features.}
        }\label{fig:training_graph}
    \end{center}
    % \vskip -0.3in
\end{figure*}

\subsection{Limitations of \benchmark{}} \label{sec:future_steps}
As any benchmark, we are fully aware that ours will not cover the entire spectrum of problems that one might be interested in. Here we summarize a few limitations that we hope to overcome in a future instantiation of this benchmark:

\textbf{Input space} We use a small $12$ dimensional observation space. This is key to achieve modest computational demand. However, richer inputs could allow for potentially more interesting forms of transfer (e.g., based on visual similarity of objects) and would allow inferring the task from the observation, which is currently impossible.

\textbf{Reliance on SAC} We use the SAC algorithm \cite{DBLP:conf/icml/HaarnojaZAL18}, which is considered a standard choice for continuous robotic tasks. However, there is a potential risk of overfitting to the particularities of this algorithm and exploring alternative RL algorithms is important.

\textbf{Task boundaries} We rely on task boundaries. One can rely on  task inference mechanisms (e.g.~\cite{DBLP:conf/nips/MilanVKBKH16, DBLP:conf/nips/RaoVRPTH19}) to resolve this limitation, though we acknowledge the importance to extend the benchmark towards allowing and testing for task inference capabilities. Also, testing for algorithms dealing with continuous distributional drift is not possible in the current format.

\textbf{Output heads} We rely on using a separate head for each new task, similar to many works on continual learning. We opt for this variant based on its simplicity and better performance than using one-hot encoding to indicate a task. We believe that the lack of semantics of the one-hot encoding would further impede transfer, as the relationship between tasks can not be inferred. We carry ablation studies with using one-hot encoding as an input and a single head architecture, a setting that is already compatible with our benchmark. We regard this aspect as an important future work, and in particular, we are exploring alternative encoding of input to make this choice more natural. A coherent domain, like \benchmark{}, provides a unique opportunity to exploit a consistent output layer as its semantics does not change between tasks.

\textbf{The difficulty and number of tasks} The number of tasks is relatively small. \cwtwenty{}, the main sequence we use, consists of only $10$ different tasks, which are then repeated. We believe the repetition of tasks is important for a CL benchmark, leading to interesting observations. We check also that results are quantitatively similar on a sequence of $30$ tasks, see Appendix \ref{sec:ablations_appendix}. However, longer sequences, potentially unbounded, are needed to understand the various limitations of existing algorithms. For example, the importance of \emph{graceful forgetting} or dealing with systems that run out of capacity, a scenario where \emph{there is no multi-task solution for the sequence of observed tasks}. This is particularly of interest for methods such as PackNet~\cite{DBLP:conf/cvpr/MallyaL18}. Additionally, we provide the number of tasks in advance. Dealing with an unknown number of tasks might raise further interesting questions. Finally, in future iterations of the benchmark, it is important to consider more complex tasks or more complex relationships between tasks to remain a challenge to existing methods. Our goal was to provide a benchmark that
is approachable by existing methods, as not to stifle progress.

\textbf{Low-level transfer} We focus on low-level transfers via neural network features and weights. As such, we do not explicitly explore the ability of the learning process to exploit the compositionality of behavior or to rely on a more interesting semantic level. While we believe such research is crucial, we argue that solving low-level transfer is equally important and might be a prerequisite. So, for now, it is beyond the scope of this work, though future iterations of the benchmark could contain such scenarios.

\section{Methods}\label{sec:cl_methods}
We now sketch $7$ CL methods evaluated on our benchmark. Some of them were developed for RL, while others were meant for the supervised learning context and required non-trivial adaptation.
%Most of them were developed in the SL context, and in some cases, non-trivial adaptation to the RL setting was required. 
We aimed to cover different families of methods; following \cite{de2019continual}, we consider three classes: regularization-based, parameter isolation and replay methods. An extended description and discussion of these methods are provided in Appendix \ref{sec:cl_methods_appendix}.

\textbf{Regularization-based Methods} This family builds on the observation that one can reduce forgetting by protecting parameters that are important for the previous tasks. The most basic approach often dubbed \textbf{L2}~\cite{DBLP:journals/corr/KirkpatrickPRVD16} simply adds a $L_2$ penalty, which regularizes the network not to stray away from the previously learned weights. In this approach, each parameter is equally important. \textbf{Elastic Weight Consolidation (EWC)}~\cite{DBLP:journals/corr/KirkpatrickPRVD16} uses the Fisher information matrix to approximate the importance of each weight.
%and propose a weighted $L_2$ penalty instead. 
\textbf{Memory-Aware Synapses (MAS)} \cite{DBLP:conf/eccv/AljundiBERT18} also utilizes a weighted penalty, but the importance is obtained by approximating the impact each parameter has on the output of the network. \textbf{Variational Continual Learning (VCL)}, follows a similar path but uses variational inference to minimize the Kullback-Leibler divergence between the current distribution of parameters (posterior) and the distribution for the previous tasks (prior).

\textbf{Parameter Isolation Methods} This family (also called modularity-based) forbids any changes to parameters that are important for the previous tasks. It may be considered as a ``hard'' equivalent of regularization-based methods. \textbf{PackNet}~\cite{DBLP:conf/cvpr/MallyaL18} ``packs'' multiple tasks into a single network by iteratively pruning, freezing, and retraining parts of the network at task change. PackNet is closely related to progressive neural networks \cite{DBLP:journals/corr/RusuRDSKKPH16},  developed in the RL context. 

\textbf{Replay Methods} Methods of this family keep some samples from the previous tasks and use them for training or as constraints to reduce forgetting. We use a \textbf{Perfect Memory} baseline, a modification of our setting which remembers all the samples from the past (i.e., without resetting the buffer at the task change). We also implemented \textbf{Averaged Gradient Episodic Memory (A-GEM)} \cite{DBLP:conf/iclr/ChaudhryRRE19}, which projects gradients from new samples as to not interfere with previous tasks. We find that A-GEM does not perform well on our benchmark. %We conjecture that stems from the fact that gradients in RL tend to be less informative.% and describe.% the problem in more detail in the appendix.

\textbf{Multi-task learning} In multi-task learning, a field closely related to CL, tasks are trained simultaneously. By its design, it does not suffer from forgetting, however, it is considered to be hard as multiple tasks ``compete for the attention of a single learning system'', see \cite{DBLP:conf/aaai/HesselSE0SH19,DBLP:journals/corr/abs-1904-11455}. We find that using reward normalization as in PopArt \cite{DBLP:conf/aaai/HesselSE0SH19} is essential to achieve good performance. See  Appendix~\ref{sec:mult_task_learning_appendix}.

\section{Experiments}
Now we present empirical results; these are evaluations of a set of $7$ representative CL methods (as described in Section \ref{sec:cl_methods}) on our Continual World benchmark.  We focus on \textit{forgetting and transfers} while keeping fixed constraints on computation, memory, number of samples, and neural network architecture. Our main empirical contributions are experiments on the long CW20 sequence and following high-level conclusions. 
%We believe they are of importance for the field. 
For a summary see Table \ref{tab:collected_resuls}, Figure \ref{fig:training_graph} and for an extensive discussion, we refer to Appendix \ref{sec:cwtwenty_graph_and_analysis_appendix} (including results for the shorter sequence, CW10). In Appendix \ref{sec:ablations_appendix} we provide various ablations and detailed analysis of sensitivity to the CL-specific hyperparameters.

% \begin{table}[t]
\begin{wraptable}{r}{0.65\linewidth}
    \vskip -0.2in
    \begin{center}
        \small
        \tabcolsep=0.11cm
\renewcommand{\arraystretch}{0.9}
\begin{tabular}{llll}
\toprule
 method   & performance       & forgetting          & f. transfer   \\
\midrule
 \textbf{Fine-tuning}        & 0.05 {\tiny [0.05, 0.06]} & 0.73 {\tiny [0.72, 0.75]}   & \textbf{0.20 {\tiny [0.17, 0.23]}}     \\
 \textbf{L2}          & 0.43 {\tiny [0.39, 0.47]} & 0.02 {\tiny [0.00, 0.03]}   & -0.71 {\tiny [-0.87, -0.57]}  \\
 \textbf{EWC}         & 0.60 {\tiny [0.57, 0.64]} & 0.02 {\tiny [-0.00, 0.05]}  & -0.17 {\tiny [-0.24, -0.11]}  \\
 \textbf{MAS}         & 0.51 {\tiny [0.49, 0.53]} & 0.00 {\tiny [-0.01, 0.02]}  & -0.52 {\tiny [-0.59, -0.47]}  \\
 \textbf{VCL}         & 0.48 {\tiny [0.46, 0.50]} & 0.01 {\tiny [-0.01, 0.02]}  & -0.49 {\tiny [-0.57, -0.42]}  \\
 \textbf{PackNet}     & \textbf{0.80 {\tiny [0.79, 0.82]}} & 0.00 {\tiny [-0.01, 0.01]} & \textbf{0.19 {\tiny [0.15, 0.23]}}     \\
 \textbf{Perfect Memory}   & 0.12 {\tiny [0.09, 0.15]} & 0.07 {\tiny [0.05, 0.10]}   & -1.34 {\tiny [-1.42, -1.27]}  \\
 \textbf{A-GEM}        & 0.07 {\tiny [0.06, 0.08]} & 0.71 {\tiny [0.70, 0.73]}   & 0.13 {\tiny [0.10, 0.16]}     \\
\midrule
 \textbf{MT}         & 0.51 {\tiny [0.48, 0.53]} & ---                 & ---                   \\
 \textbf{MT (PopArt)}  & 0.65 {\tiny [0.63, 0.67]} & ---                 & ---                   \\
            \midrule
            \textbf{RT}          & ---                                & ---                         & \textbf{0.46}                             \\
            \bottomrule
        \end{tabular}
        \caption{\small Results on \cwtwenty{}, for CL methods and multi-task training.  Metrics are defined in Section \ref{sec:metrics}, RT is eq. \eqref{eq:forward_transfer_upperbound}. We used $20$ seeds and provide 90\% confidence intervals. } %\michal{I wonder if these refs to equations are really helpful, especially since the distance in text is big... Maybe instead write "performance, forgetting, tranfsfer as defined in Section X?} 
        \label{tab:collected_resuls}
    \end{center}
    \vskip -0.3in
\end{wraptable}
% \end{table}

\textbf{Performance} The performance (success rate) averaged over tasks (eq. \eqref{eq:performance}) is a typical metric for the CL setting. PackNet seems to outperform other methods, approaching $0.8$ from the maximum of $1.0$, outperforming multi-task solutions which might struggle with different reward scales, a problem elegantly avoided in the CL framing. Other methods perform considerably worse. A-GEM and Perfect Memory struggle. We further discuss possible reasons in Section~\ref{sec:rl_algorithmic_challenges}.
%
%is perhaps the single most important metric of CL performance. We recall that we tuned hyperparameters to maximize its value. PackNet is a clear winner, approaching $0.8$. Other methods perform significantly worse. A-GEM and Perfect Memory struggle, we conjecture possible explanations in Section \ref{sec:rl_algorithmic_challenges}.\piotrms{Final: This section is slightly awkward}%, but it would be more (?) awkward not to say anything about performance.}

\textbf{Forgetting} We observe that most CL methods are usually efficient in mitigating forgetting. However, we did not notice any boost when revisiting a task (see Figure~\ref{fig:training_graph}). Even if a different output head was employed, relearning the internal representation should have had an impact unless it changed considerably when revisiting the task. Additionally, we found A-GEM difficult to tune; consequently, with the best hyperparameter settings, it is relatively similar to the baseline  fine-tuning method (see details in Appendix \ref{sec:algorithmic_and_conceptual_challenges_appendix}).

\textbf{Transfers} %The average forward transfer, as defined in \eqref{eq:forward_transfer}, for all methods and CW20 is summarized in Table \ref{tab:collected_resuls} and Appendix \ref{sec:forward_transfer_appendix} (Table \ref{tab:long_forward_transfer_table} and Figures \ref{fig:transfer_all}, \ref{fig:transfer_first_10}, \ref{fig:transfer_second_10}).
For all methods, forward transfer for the second ten tasks (and the same tasks are revisited) drops compared to the first ten tasks. This is in stark contrast to forgetting, which seems to be well under control. Among all methods, only fine-tuning and PackNet are able to achieve positive forward transfer ($0.20$ and $0.19$, resp.) as well as on the first ($0.32$ and $0.21$, resp.) and the second ($0.08$ and $0.17$, resp.) half of tasks. However, these are considerably smaller than $\text{RT}=0.46$, which in principle can even be exceeded, and which should be reached by a model that remembers all meaningful aspects of previously seen tasks, see \eqref{eq:forward_transfer_upperbound}. These results paint a fairly grim picture: we would expect improvement, rather than deterioration in performance, when revisiting previously seen tasks. There could be multiple reasons for this state of affairs. It could be attributed to the loss of plasticity,
similar to the effect observed in~\cite{DBLP:conf/nips/AshA20}.
%and a somewhat similar effect was observed in the context of supervised learning, see e.g., \cite{DBLP:conf/nips/AshA20}. 
Another reason could be related to the interference between CL mechanisms or setting and RL, for instance, hindering exploration. We did not observe any substantial cases of \textbf{backward transfer}, even though the benchmark is well suited to study this question due to the revisiting of tasks. See Appendix~\ref{sec:forgetting_appendix}.

%For details see Section \ref{sec:forward_transfer_appendix}. \piotrm{Forward transfers are rather unsatisfactory, especially for longer sequences.}
\begin{wrapfigure}{l}{0.70\linewidth}
    %\vskip -0.in
    \begin{center}
        \centerline{\includegraphics[width=0.7\columnwidth]{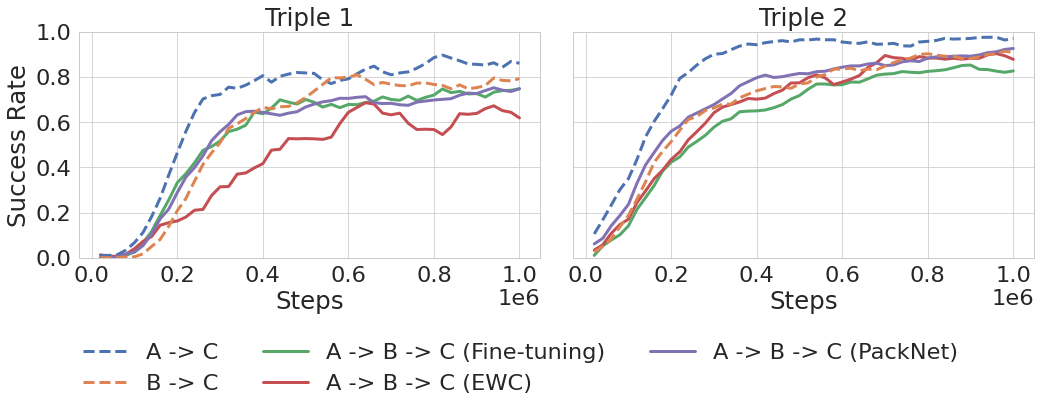}}
        \caption{\small \textbf{How forgetting impacts the forward transfer.} %The figure presents success rate throughout training on two sequences of three tasks. 
        Two different triplets of tasks learnt in sequence. An ideal agent learning on a sequence $A \rightarrow B \rightarrow C$ should have at least as good performance on task $C$ as an agent which just learns $A \rightarrow C$. In reality, an interfering task $B$ reduces this transfer, even when continual learning approaches are used.  % Right plot showcases that this effect appears even when learning on $A \rightarrow B \rightarrow A$.
        }
        \label{fig:triplets}
    \end{center}
    \vskip -0.3in
\end{wrapfigure}

\textbf{Triplets experiments} We illustrate how forgetting and forward transfer interact with each other in a simpler setting of three task sequences, see Figure~\ref{fig:triplets} and Appendix~\ref{sec:triplets_appendix}.
%Figure \ref{fig:triplets} thus shows a situation where remembering might be required to improve forward transfer and the overall training speed. 
We focus on sequences of tasks $A \rightarrow B \rightarrow C$, where $A \rightarrow C$ has significant positive forward transfer and $B \rightarrow C$ has a smaller or even negative transfer.
An efficient CL agent should be able to use information from $A$ to get good performance on $C$. However, interference introduced by $B$ reduces the final forward transfer (see Figure~\ref{fig:triplets}). The drive for reducing forgetting in CL agents has been primarily to perform well on previous tasks when we revisit them. With this example, we argue that an equally important reason to improve the memory of CL agents is to efficiently use past experiences to learn faster on new tasks. Currently, the tested CL methods often are not able to outperform the forgetful fine-tuning baseline. Observe that even the modularity-based PackNet approach struggles with this task. This possibly indicates that using the activation mask from task $B$ is enough to deteriorate the performance.

\textbf{PackNet} PackNet stands out in our evaluations.
We conjecture that developing related methods might be a promising research direction.
Besides further increasing performance, one could mitigate the limitations of PackNet.
%\michal{Replace last two sentences with: "However, it comes with a set of limitations; therefore we believe that developing less constrained related methods could be a promising research direction.}
PackNet relies on knowing task identity during evaluation. While this assumption is met in our benchmark, it is an interesting topic for future research to develop methods that cope without task identity. Another nuisance is that PackNet assigns some fixed fraction of parameters to a task. This necessitates knowledge of the length of the sequence in advance.
Additionally, when the second ten tasks of \cwtwenty{} start, PackNet performance degrades, showing its potentially inefficient use of capacity and past knowledge, given that the second ten tasks are identical with the first ten and hence no additional capacity is needed. %Finally, by construction PackNet does not allow backward transfer, though no other method had exhibited backward transfer either. 
In a broader context, we speculate that parameter isolation methods might be a promising direction towards better CL methods.

\textbf{Resources usage} In practical applications it is important to consider resources usage. All tested methods have relatively small overheads. For example, PackNet needs only $15\%$ more time than the baseline fine-tuning and it requires $50\%$ more neural network parameters (which is negligible when small networks like ours). See Appendix \ref{sec:resources_comparison} for details concerning other methods.

\textbf{Other observations} In stark contrast with the supervised learning setting, we found that replay based methods (Perfect Memory and A-GEM) suffer from poor performance. This is even though we allow for a generous replay, which could store the whole experience. Explaining and amending this situation is, in our view, an important research question. We conjecture that this happens due to the regularization of the critic network (which was unavoidable for these methods). We found multi-task learning attaining lower scores than PackNet, the best CL method and comparable to the second one, EWC. We think this suggests interesting research directions for multi-task learning.

% Finally, we refer the reader to Appendix~\ref{sec:cwtwenty_graph_and_analysis_appendix} for further analysis and ablation studies providing a more complete picture of the benchmark which for brevity constraints we could not fully convey here. % \razp{maybe re-edit but something of this sort at the end.}

\subsection{RL-Related Challenges}\label{sec:rl_algorithmic_challenges}

% \maciejws{Mźaybe we could move this to the beginning of this section, before the Forgetting paragraph?}
Reinforcement learning brings a set of issues not present in the supervised learning setting, e.g., exploration, varying reward scales, and stochasticity of environments. We argue that it is imperative to have a reliable benchmark to assess the efficiency of CL algorithms with respect to these problems. We find that some current methods are not well adjusted to the RL setting and require non-trivial conceptual considerations and careful tuning of hyperparameters, see details in Appendix~\ref{sec:algorithmic_and_conceptual_challenges_appendix}. 

An important design choice is whether or not to regularize the critic in the actor-critic framework (e.g. in SAC). We find it beneficial to focus on reducing forgetting in the actor while allowing the critic to freely adapt to the current task (note that critic is used only in training of the current task), similar to \cite{DBLP:conf/icml/Schwarz0LGTPH18}. %In fact, improvements offered by this approach are so significant that some of the tested methods are able to outperform the results obtained in the multi-task setting, where critic has to ''cover'' all tasks at the same time. 
% This is much different from the supervised setting, in which multi-task learning is often an upper bound to CL.
On the other hand, a forgetful critic is controversial. This can be sharply seen when the same task is repeated and the critic needs to learn from scratch. Additionally, not all methods can be trivially adapted to the 'actor-only regularization' setting, as for example replay based methods. In Appendix \ref{sec:algorithmic_and_conceptual_challenges_appendix} we examine these issues empirically, by showing experiments with critic regularization for EWC.

Another aspect is the exploration and its non-trivial impact on transfers. As it was observed, transfers from a given task to the same one are sometimes poor. We show in Appendix \ref{sec:diagonal_transfers_appendix} that this results from the fact that at the task change the replay buffer is emptied and SAC collects new samples from scratch, usually by using the uniform policy. Learning on these random samples reduces performance on the current task and thus the forward transfer. Experimentally, we find that  not resetting the buffer or using the current policy for exploration improves the transfer on the diagonal. % At the same time, it could lead to d off-diagonal~transfers.  %\maciejw{Additional point: the samples are (extremely) off-policy}

\section{Conclusions and Future Work}\label{sec:conclusions_and_futher_work}
In this work, we present \benchmark{}, a continual reinforcement learning benchmark, and an in-depth analysis of how existing methods perform on it. The benchmark is aimed at facilitating and standardizing the CL system evaluation, and as such, is released with code, including implementation of $7$ representative CL algorithms. We argue for more attention to \emph{forward transfer} and the interaction between forgetting and transfer, as many existing methods seem to sacrifice transfer to alleviate forgetting. In our opinion, this should not be the aim of CL, and we need to strike a different balance between these objectives.

We made several observations, both conceptual and empirical, which open future research directions. In particular, we conjecture that parameter isolation methods are a promising direction. Further, we identified a set of critical issues at the intersection of RL and CL. Resolving critic regularization and efficient use of multi-task replays seem to be the most pressing ones. Our benchmark highlights some challenges, which in our view are relevant and tangible now. In the long horizon, achieving high-level transfers, removing task boundaries, and scaling up are among significant goals for future editions of \benchmark{}. Our work is foundational research and does not lead to any direct negative applications.

\begin{ack}
We would like to thank Stanisław Jastrzębski for stimulating talks and help while preparing the manuscript. The work of PM was supported by the Polish National Science Center grant UMO-2017/26/E/ST6/00622. The work of MW was funded by Foundation for Polish Science (grant no POIR.04.04\allowbreak.00-00-14DE/18-00 carried out within the Team-Net program co-financed by the European Union under the European Regional Development Fund. This research was supported by the PL-Grid Infrastructure. Our experiments were managed using \url{https://neptune.ai}. We would like to thank the Neptune team for providing us access to the team version and technical support.
\end{ack}

\appendix

% \section{Appendix}
\newpage
\section{Technical details}\label{sec:technical_details}
\subsection{Tasks details}
\label{sec:obs_act_rew}
\benchmark{} consists of a series of robotic tasks performed by a simulated Sawyer robot, each lasting for $200$ steps. Individual tasks come from \metaworld{}~\cite{metaworld} (benchmark released on MIT license) and as such, they have a significantly shared structure. This includes a common definition of the observation and action space as well as a similar reward structure.% \michal{Actually, the same reward structure, right?}
\paragraph{Observations} The observation space is $12$-dimensional, consisting of:
\begin{itemize}
    \item the $3$D Cartesian position of the robot end-effector ($3$ numbers),
    \item the $3$D Cartesian positions of one or two objects that the robot manipulates\footnotemark{} (6 numbers),
    \item the $3$D Cartesian position of the goal to be reached (3 numbers).
\end{itemize}
\footnotetext{For tasks with one object, the excessive coordinates are zeroed out.}
\paragraph{Actions} The action space is $4$-dimensional, including the requested $3$D end-effector position change and the gripper actuator delta. %\piotrm{someone verify please} \michal{I think this is ok.}

\paragraph{Rewards} \metaworld{} proposes a careful design of rewards, which ensures that tasks are solvable and exhibit shared structure across the tasks. These rewards are task-specific, see \cite[Section 4.2, Table 3]{metaworld} and are used only in training.

In the evaluation phase, a binary success metric is used
\begin{equation}
    \mathbb{I}_{{\| o-g \|}_{2}}< \varepsilon, \label{eq:evaluation_metric}
\end{equation}
where $o,g$ are the $3$D positions the object and goal, respectively, and  $\varepsilon$ is a task-specific threshold, see \cite[Table 4]{metaworld}.

\subsection{Sequences}\label{sec:sequences_appendix}
Our benchmark consists of two long CW10, \cwtwenty{} sequences and $8$ triples.

The CW10 sequence is: {\small \textsc{hammer-v1}, \textsc{push-wall-v1}, \textsc{faucet-close-v1}, \textsc{push-back-v1}, \textsc{stick-pull-v1},
\textsc{handle-press-side-v1}, \textsc{push-v1}, \textsc{shelf-place-v1}, \textsc{window-close-v1}, \textsc{peg-unplug-side-v1}}.

The \cwtwenty{} is CW10 repeated twice. The triples are presented in Section \ref{sec:triplets_appendix}.

\subsection{Training and evaluation details}\label{app:training_details}
We use an implementation of SAC algorithm \cite{DBLP:conf/icml/HaarnojaZAL18, DBLP:journals/corr/abs-1812-05905} based on \cite{SpinningUp2018} , which we ported to TensorFlow 2. SAC is an actor-critic type algorithm that uses neural networks in two different roles: the policy network (actor) and the value function network (critic).

Both the actor and critic are randomly initialized at the start of the first task and then are trained through the whole sequence, retaining parameters from previous tasks.

In all cases, except for A-GEM and Perfect Memory, we use CL algorithms only for the policy networks, for details see Section \ref{sec:critic_regularizaiton_appendix}.

\paragraph{Task handling} Each task is trained for $1$M steps. The replay buffer is emptied when a new task starts. For the first $10$K steps, actions are sampled uniformly from the action space. After these, actions are sampled using the current policy. We use a warm-up period of $1$K steps before training neural networks.

\paragraph{SAC training} We perform $50$ optimization steps per each $50$ interactions collected. To this end, we use the Adam optimizer \cite{DBLP:journals/corr/KingmaB14}. Its internal statistics are reset for each task. The maximum entropy coefficient $\alpha$ is tuned automatically \cite{DBLP:journals/corr/abs-1812-05905}, so that the average standard deviation of Gaussian policy matches the target value $\sigma_t =0.089$.

\paragraph{Policy evaluation} Performance on all tasks is measured (mean of \eqref{eq:evaluation_metric} on $10$ stochastic trajectories) every $20K$ steps. In the case of some metrics, we average over $5$ points in time to get smoother results.

% These evaluations are used to compute the following metrics: \maciejw{is something missing here?} for forward transfer, all the points are used; for average performance and forgetting last 5 points from each task are used (to smoothen the metrics).

\subsection{Network architecture}\label{sec:network_architecture_appendix}
We use a network with $4$ linear layers, each consisting of $256$ neurons. We add Layer Normalization \cite{DBLP:journals/corr/BaKH16} after the first layer. We use leaky ReLU activations (with $\alpha = 0.2$) after every layer except the first one, where we use the tanh activation, which is known to work well with Layer Normalization layers.

The network has as many heads as the corresponding sequence, e.g. $20$ for \cwtwenty{}. In Appendix~\ref{sec:onehot} we present experiments in which a single head is used, and the task id input is provided.

The presented architecture is used for both the actor and the critic. The actor network input consists of an observation, and it outputs the mean and log standard deviation of the Gaussian distribution on the action space. These outputs are produced for each task by its respective head. The critic network input consists of an observation and action, and it outputs a scalar value for each task head. %\piotrm{Do we have multiple heads for the critic? These are not needed in most cases (though not harmuful).} \michal{Yes, we use it}

% \piotrm{Write in little more details. I.e. say something about the value network and policy network.}

\subsection{Hyperparameters} \label{sec:hyperparameters_appendix}
We performed hyperparameter search in two stages. In the first stage, we selected hyperparameters of the SAC algorithm, which are common among all the methods. We measured the performance training on the \cwtwenty{} sequence using only the fine-tuning method. As the metric, we used the mean performance at the end of the training of each task. We present the search space and selected parameters in Table~\ref{hparams:general}.

%  First, we select the base hyperparameters that are common among all the methods, i.e.\ learning rate, batch size, discount factor $\gamma$ and target output standard deviation (see Appendix~\ref{app:training_details}). We perform the search by checking how respective settings perform with fine-tuning method. The criterion for selecting the best parameters is averaged performance, where for every task we take the performance at the end of learning of this task. This is because regular average performance for the final model is poor for the fine-tuning method no matter what parameters, and we want to select such a setting in which at least single-task learning is efficient.
In the second stage, we tuned method-specific hyperparameters using the final average performance as the objective. These are presented in Appendix~\ref{sec:cl_methods_appendix}.

\begin{table}
    \centering
    \begin{tabular}{ lcc }
        \toprule
        parameter                    & search space                                         & selected value \\
        \midrule
        learning rate                & $\{\num{3e-5}, \num{1e-4}, \num{3e-4}, \num{1e-3}\}$ & $\num{1e-3}$   \\
        batch size                   & $\{128, 256, 512\}$                                  & $128$          \\
        discount factor $\gamma$     & $\{0.95, 0.99, 0.995\}$                              & $0.99$         \\
        target output std $\sigma_t$ & $\{0.03, 0.089, 0.3\}$                               & $0.089$        \\
        replay buffer size           & ---                                                  & 1M             \\
        \bottomrule
    \end{tabular}
    \caption{Search space for the common hyperparameters.}
    \label{hparams:general}
\end{table}

\subsection{Bootstrap confidence intervals}\label{sec:confidence_intervals}
We use non-parametric bootstrap \cite{efron1994introduction} to calculate confidence intervals. Reinforcement learning typically has rather high variability. We found that using $20$ seeds usually results in informative intervals.

Unless specified otherwise, in all numerical results below, we use at least $20$ seeds and report $90\%$ confidence intervals.

\subsection{Infrastructure used} \label{sec:infrastructure_used} The typical configuration of a computational node used in our experiments was: the Intel Xeon E5-2697 $2.60$GHz processor with $128$GB memory. On a single node, we ran $3$ or $4$ experiments at the same time. A single experiment on the CW20 sequence takes about $100$ hours (with substantial variability with respect to the method used). We did not use GPUs; we found that with the relatively small size of the network, see Section \ref{sec:network_architecture_appendix} it offers only slight wall-time improvement while generating substantial additional costs. 

\section{CL methods}\label{sec:cl_methods_appendix}
We now describe seven CL methods evaluated on our benchmark: L2, EWC, MAS, VCL, PackNet, Perfect Memory and Reservoir Sampling. Most of them were developed in the supervised learning context and in many cases non-trivial adaptation to RL setting was required, which we describe in Appendix \ref{sec:algorithmic_and_conceptual_challenges_appendix}. We picked the following approaches in an attempt to cover different families of methods considered in the community. Namely, following \cite{de2019continual}, we consider three classes: regularization-based methods, parameter isolation methods, and replay methods.

\subsection{Regularization-based methods}
Regularization-based methods build on an observation that connectionist models, like neural networks, are heavily overparametized. A relatively small number of parameters is genuinely relevant for the previous tasks. Thus, if ``protected'', forgetting might be mitigated. Regularization methods implement this by an additional regularization loss, often in the $l_2$ form. For example, in order to remember task $1$ while learning on task $2$ we add a regularization cost of the form:
\begin{equation}
    \lambda \sum_k F_k \left(\theta_k - \theta_k^1\right)^2,\label{eq:generic_regularization}
\end{equation}
where $\theta_k$ are the neural network current weights, $\theta^1_k$ are the values of weights after learning the first task, and $\lambda>0$ is the overall regularization strength. The coefficients $F_k$ are critical, as they specify how important parameters are. Regularization-based methods are often derived from a Bayesian perspective, where the regularization term enforces the learning process to stay close to a prior, which now incorporates the knowledge acquired in the previous tasks.

There are multiple ways to extend this approach to multiple tasks. Arguably, the simplest one is to keep the importance vector $F^i$ and parameters $\theta^i$ for each task $i$. This, however, suffers from an unacceptable increase in memory consumption. We thus opt for a memory-efficiency solution presented in  \cite{huszar2018note}. Namely, we store the sum of importance vectors and the weights after training the last task. Formally, when learning $m$-th task, the penalty has the following form:
\begin{equation}
    \lambda \sum_k \Big(\sum_{i=1}^{m-1} F_k^i\Big) \left(\theta_k - \theta_k^{m-1}\right)^2.\label{eq:generic_regularization_cumulative}
\end{equation}

\paragraph{L2} This method is used as a baseline in \cite{DBLP:journals/corr/KirkpatrickPRVD16}. It declares all parameters to be equally important, i.e. $F^i_k = 1$ for all $k, i$. Despite its simplicity, it is able to reduce forgetting, however, often at the cost of a substantial reduction in the ability to learn.

For the L2 method, we tested the following hyperparameter values: $\lambda \in \{ 10^{-2}, 10^{-1}, 10^{0}, 10^1, 10^2, 10^3, 10^4, 10^5 \}$; selected value is $10^5$.

\paragraph{EWC} Elastic Weight Consolidation \cite{DBLP:journals/corr/KirkpatrickPRVD16} adheres to the Bayesian perspective. Specifically, it proposes that the  probability distribution of parameters of the previous tasks is a prior when learning the next task. Since this distribution is intractable, it is approximated using the diagonal of the Fisher Information Matrix. Namely, $F_k = \mathbb{E}_{x\sim \mathcal D}\mathbb{E}_{y\sim p_{\theta}(\cdot | x)} \Big(\nabla_{\theta_k} \log p_{\theta_k}(y | x) \Big)^2$. The outer expectation is approximated with a sample of $2560$ examples from the replay buffer $\mathcal{D}$. The inner expectation can be calculated analytically for Gaussian distributions used in this work, see Section \ref{sec:ewc_fisher} for details. We clipped the calculated values $F_k$ from below so that the minimal value is $10^{-5}$.

For the EWC method we tested the following hyperparameter values: $\lambda \in \{ 10^{-2}, 10^{-1}, 10^{0}, 10^1, 10^2, 10^3, 10^4, 10^5 \}$; selected value is $10^4$.

% \paragraph{MAS} Memory Aware Synapses, \cite{DBLP:conf/eccv/AljundiBERT18}, continually aggregates the importance of each neural network weight by measuring the sensitivity of the output with respect to the weigh perturbations. \piotrm{possibly put formula for the gradient, 2021-01-30}
%Given a new sample which is fed to the network, MAS accumulates an importance measure for each parameter of the network, based on how sensitive the predicted output function is to a change in this parameter

\paragraph{MAS} Memory Aware Synapses \cite{DBLP:conf/eccv/AljundiBERT18} estimates the importance of each neural network weight by measuring the sensitivity of the output with respect to the weight perturbations. Formally,  $F_k = \mathbb{E}_{x\sim \mathcal{D}} \Big ( \frac{\partial [\|g(x)\|_2^2}{\partial \theta_k} \Big )$, where $g$ is the output of the model and expectation is with respect to the data distribution $\mathcal{D}$. To approximate the expected value in the formula above, we sampled $2560$ examples from the buffer.

For the MAS method we tested the following hyperparameter values: $\lambda \in \{ 10^{-2}, 10^{-1}, 10^{0}, 10^1, 10^2, 10^3, 10^4, 10^5 \}$; selected value is $10^4$.

\paragraph{VCL} Variational Continual Learning \cite{DBLP:conf/iclr/NguyenLBT18} builds on the Bayesian neural network framework. Specifically, it maintains a factorized Gaussian distribution over the network parameters and applies the variational inference to approximate the Bayes rule update. Thus, the training objective contains an additional loss component
\begin{equation}
    \lambda D_{\text{KL}}(\theta \parallel \theta^{m-1}),
\end{equation}
where $D_{\text{KL}}$ denotes the Kullback-Leibler divergence and $\theta^{m-1}$ is the parameter distribution corresponding to the previous tasks. This loss can be viewed as regularization term similar to \eqref{eq:generic_regularization}.

We introduced multiple changes to make VCL usable in the RL setting. First, we tune $\lambda$. The original work sets  $\lambda = 1/N$, where $N$ is the number of samples in the current task, which we found performing poorly. Second, we discovered that not using prior (equivalently, setting $\lambda=0$) on the first task yields much better overall results. Third, along with a standard smoothing procedure, \cite{DBLP:conf/icml/BlundellCKW15} averages the prediction using $10$ samples of the weights. We opted for taking $1$ sample, since it  performed comparably with $10$ samples, but significantly decreased the training time. Fourth, we do not use the coresets mechanism due to a high computational cost. Further discussion of these issues is presented in Section \ref{sec:vcl_problems_appendix}.

For the VCL method we initialize parameters with $\mathcal{N}(\mu, \sigma)$, $\mu=0,\sigma = 0.025$. We tested the following hyperparameter values: $\lambda \in \{ 10^{-7}, 10^{-6}, 10^{-5}, 10^{-4}, 10^{-3}, 10^{-2}, 10^{-1}, 1 \}$; selected value is $1$.

\subsection{Parameter isolation methods}

Parameter isolation methods (also called modularity-based methods) retain already acquired knowledge by keeping certain ranges of parameters fixed. Such an approach may be considered as a hard constraint, as opposed to the soft penalties of regularization-based methods.
% employ the view that certain ranges of parameters should be kept fixed to re

% The second family of CL methods use parameter isolation or modularity. These methods completely forbid changes to parameters which were decided to be important for the previous tasks. It may be thus seen as a hard constraint, opposed to the soft penalty-based constraints present in regularization-based methods.

\paragraph{PackNet} Introduced in \cite{DBLP:conf/cvpr/MallyaL18}, this method iteratively applies pruning techniques after each task is trained. In this way, it ``packs the task'' into a subset of the neural network parameters leaving others available for the subsequent tasks; the parameters associated with the previous tasks are frozen. Note that by design, this method completely avoids forgetting.
Compared to earlier works, such as progressive networks~\cite{DBLP:journals/corr/RusuRDSKKPH16}, the model size stays fixed through learning. However, with every task, the number of available parameters shrinks.

Pruning is a two-stage process. In the first one, a fixed-sized subset of parameters most important for the task is chosen. The size is set to  $5 \%$ in the case of \cwtwenty{}. In the second stage, the network spanned by this subset is fine-tuned for a certain number of steps. In this process, we only use the already collected data.

All biases and normalization parameters (of layer normalization) are not handled by PackNet. They remain frozen after training the first task. Also, the last layer is not managed by PackNet, as each task has its own head.

In PackNet we used global gradient norm clipping \num{2e-5}, as explained in Section \ref{sec:pack_net_appendix_problems}.

For PackNet, we tested the following hyperparameter values: number of fine-tuning steps in $\{50K, 100K, 200K\}$; selected value if $100K$. %\piotrm{I make this uniform with other sections. } \maciejw{MZ: I would add the gradient clipping hps here as well.}

\subsection{Replay methods}
\label{sec:replay_methods_appendix}
Replay methods maintain a buffer of samples from the previous tasks and use them to reduce forgetting. These are somewhat similar but should not be confused with the replay buffer technique commonly used by off-policy RL algorithms (including SAC).

\paragraph{Reservoir Sampling} Reservoir sampling \cite{DBLP:journals/toms/Vitter85} is an algorithm, which manages the buffer so that its distribution approximates the empirical distribution of observed samples. This method is suitable for CL and often used as a standard mechanism for gathering samples \cite{DBLP:conf/iclr/ChaudhryRRE19}. In our implementation, we augment samples with the task id and store them in the SAC replay buffer. For the sake of simplicity, in the main part of the paper we stick to an otherwise unrealistic assumption of the unlimited capacity buffer, which we dubbed as \textbf{Perfect Memory}.

We find that it does not work very well, even if we increase the number of samples replayed at each step (batch size) to $512$, which significantly extends the training time.

For the reservoir sampling methods, we tested the following hyperparameter values: $\text{batch size} \in \{ 64, 128, 256, 512 \}$ (selected value $=512$), $\text{replay size} \in \{ 5 \cdot 10^6, 1 \cdot 10^7, 2 \cdot 10^7 \}$ (selected value $=2 \cdot 10^7$). The value $2\cdot 10^7$ corresponds to Perfect Memory, as it is able to store all experience of CW20.

% \paragraph{Perfect Memory} The simplest possible approach is to store all the possible examples from the past. However, we find that it does not work very well, even if we increase the batch size to $512$. \piotrm{It is unclear what batchsize is.}

% For the Perfect Memory methods, we tested the following hyperparameter values: $\text{batch size} \in \{ 128, 256, 512 \}$.

% \paragraph{Reservoir Sampling}

% In the main paper we only considered the setting where we are able to store all examples. However in real-life situations with memory constraints this may be infeasible. Here, we consider a situation when the size of the replay buffer is much smaller, e.g. is only able to hold one task, as in our basic setup. This necessitates using a special strategy of sampling to the buffer. Standard first-in-first-out (FIFO) approach would mean that samples from the previous tasks would be quickly discarded.

% Reservoir sampling \cite{DBLP:journals/toms/Vitter85} is an algorithm, which manages the buffer so that its distribution approximates the empirical distribution of observed samples. This method is suitable for CL and often used as a standard mechanism for gathering samples \cite{DBLP:conf/iclr/ChaudhryRRE19}. We implement this approach simply by accumulating in the SAC replay buffer transitions of the previous tasks. Note that with unlimited buffer size reservoir sampling is the same as perfect memory. In section \ref{sec:rsampling_appendix} we show that this approach performs badly.

\paragraph{A-GEM} Averaged Gradient Episodic Memory \cite{DBLP:conf/iclr/ChaudhryRRE19} frames continual learning as a constrained optimization problem. Formally, a loss for the current task $\ell(\theta, \mathcal{D})$ is to be minimized under the condition that the losses on the previous tasks are bounded $\ell(\theta, \mathcal{M}_k) \leq \ell_k$, where $\ell_k$ is the previously observed minimum and $\mathcal{M}_k$ are the buffer samples for task $k$. Unfortunately, such constraint is not tractable for neural networks. \cite[Section 4]{DBLP:conf/iclr/ChaudhryRRE19} proposes an approximation based on the first-order Taylor approximation:
\begin{equation}
    \label{eq:agem_approx}
    \langle \nabla_\theta \ell(\theta, \mathcal{B}_{new}), \nabla_\theta \ell(\theta, \mathcal{B}_{old}) \rangle > 0,
\end{equation}
where $\mathcal{B}_{new}, \mathcal{B}_{old}$ are, respectively, batches of data for the current and previous tasks. This constraint can be easily implemented by a gradient projection: %\kucil{the constraint in (9) is sharp}

\begin{equation}\label{eq:agem_projection}
    \nabla_\theta \ell(\theta, \mathcal{B}_{new})
    - \frac{\left \langle \nabla_\theta \ell(\theta, \mathcal{B}_{new}), \nabla_\theta \ell(\theta, \mathcal{B}_{old}) \right \rangle}
    {\left \langle \nabla_\theta \ell(\theta, \mathcal{B}_{old}), \nabla_\theta \ell(\theta, \mathcal{B}_{old}) \right \rangle}
    \nabla_\theta \ell(\theta, \mathcal{B}_{old})
\end{equation}

For A-GEM, we set episodic memory per task to 10K. We tested the following hyperparameter values: $\text{episodic batch size} \in \{ 128, 256 \}$ (selected value $=128$).

\subsection{Resources comparison} \label{sec:resources_comparison}

As we highlight in Section \ref{sec:continual_learning_definitions}, continual learning is characterized by multiple, often conflicting, desiderata. In practice, except for the metrics studied in the paper (see Section \ref{sec:metrics}), one could pay attention to the efficiency in terms of resource usage. To this end, we check how the tested methods perform in terms of computational power, measured by average wall time and memory efficiency, measured by the overhead in the number of parameters, and the number of examples in the buffer. The results presented in Table \ref{tab:efficiency} show that most of the methods use a fairly modest amount of resources, exceptions being reservoir sampling, which needs twice as much computation time as the baseline fine-tuning, and VCL, which requires more parameters due to using Bayesian neural networks.

\begin{table}[h!]
\begin{tabular}{lrp{4.5cm}r}
\toprule
Method  & Normalized wall time & Normalized parameter overhead & Examples to remember \\
\midrule
Fine-tuning &           1.00 &  1.0 & {\small N/A} \\
Reservoir &          2.20 & 1.0 & 20M \\
A-GEM &          1.47 & 1.0 & 1M \\
EWC &          1.13 & 2.0 {\small (parameters of the previous task \& importance weights)} & {\small N/A} \\
L2 &          1.12 & 1.5 {\small(parameters of the previous task)} & {\small N/A} \\
MAS &          1.13 & 2.0 {\small(parameters of the previous task \& importance weights)} & {\small N/A} \\
PackNet &      1.15 & 1.5(*) {\small(mask assigning parameters to task)} & {\small N/A} \\
VCL &          1.28 & 4.0 {\small(parameters of the previous task \& network is two times bigger as we model standard deviation)}   & {\small N/A} \\
\bottomrule
\end{tabular}
\caption{Results for computational and memory overhead for each method (normalized with respect to Fine-tuning). (*)Masks in PackNet (integers) can be encoded with fewer bits than the parameters (floats).} 
\label{tab:efficiency}
\end{table}

\section{Algorithmic and conceptual challenges of the RL+CL setting}\label{sec:algorithmic_and_conceptual_challenges_appendix}
% \piotrm{Change the title. Any suggestions?} "Adaptations to CL setting ???"

This section lists various challenges that we encountered while implementing the CL methods described in Section \ref{sec:cl_methods_appendix}. In most cases, they stem from the fact that the methods were initially developed for supervised learning and needed to be adopted to the more algorithmically complex RL setting.

\subsection{Critic regularization} \label{sec:critic_regularizaiton_appendix}

\begin{figure}
    \centering
    \includegraphics[width=\textwidth]{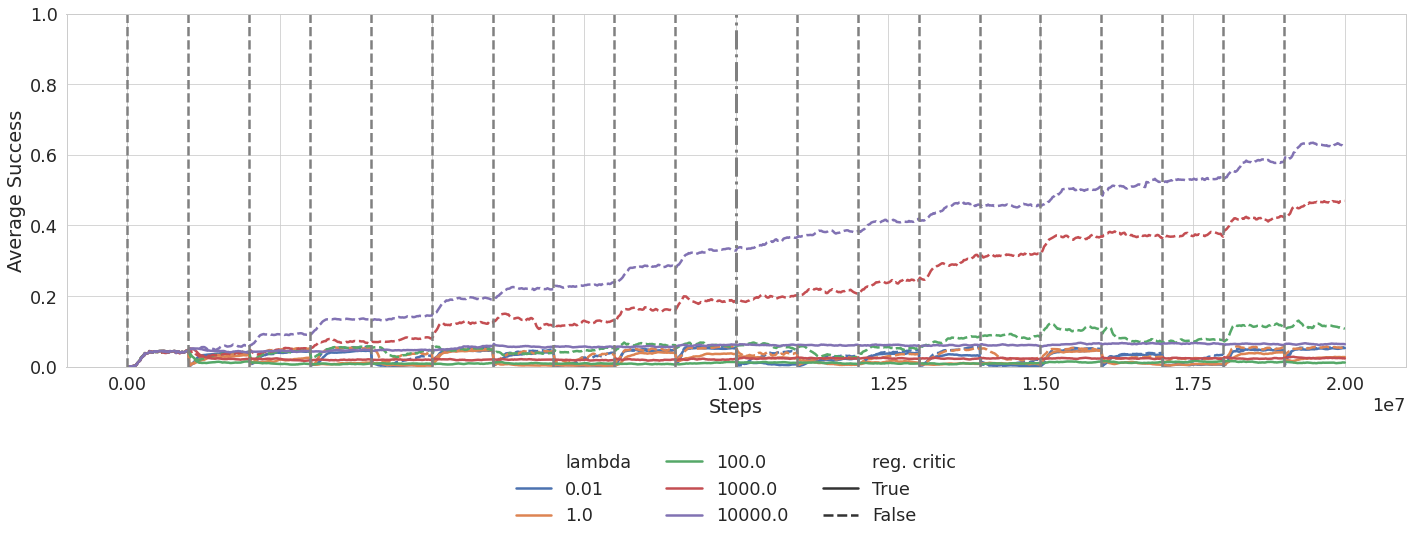}
    \caption{Average performance for EWC with various settings of the critic regularization, see \eqref{eq:generic_regularization}. The reference dashed lines show the performance without critic regularization.}
    \label{fig:ewc_critic_ablation_appendix}
\end{figure}

One of perhaps most important algorithmic conclusions is that \emph{regularizing the critic network is harmful to performance}. We recall that in all methods, except for AGEM and Perfect Memory, the critic is only used during the training. Consequently, it needs to store only the value function for the current task, and thus forgetting is not a (primary) issue. Further, one can opt for not using any CL method on the critic network. The rationale for such a decision is to maintain the full network ``plasticity'' (which is typically reduced by CL methods). In our experiments, we show that this is the case. However, we stress that not regularizing critic might be controversial. A  suggestive straightforward example, present in the \cwtwenty{} sequence, is a situation when a task is repeated after a while, and the critic needs to be trained from scratch. %\maciejw{Maybe we should comment that there is no observed improvement in our case?}

We now present an in-depth analysis using EWC as an example. To this end, we performed a separate hyperparameter search for this setting for the regularization parameter $\lambda$ in \eqref{eq:generic_regularization}. Results, as presented in Figure \ref{fig:ewc_critic_ablation_appendix}, clearly indicate that the regularized cases struggle to achieve a decent result, in stark contrast to the non-regularized references (dashed lines) (we tested more values of $\lambda$, which are not included for better readability).% \maciejw{We tested more values of lambda and plot half of them for better readability.}

We consider critic regularization issues to be an important research direction.

\subsection{EWC Fisher matrix derivation}
\label{sec:ewc_fisher}
For the sake of completeness, we provide an analytical derivation of the Fisher matrix diagonal coefficients in the case of Gaussian distribution used by the SAC policy. In experiments presented in Section \ref{sec:critic_regularizaiton_appendix} we use these formulas also for the critic network assuming a normal distribution with constant $\sigma^2 = 1$.

\begin{fact}
    Let $\mu:\mathbb{R} \mapsto \mathbb{R}, \sigma:\mathbb{R} \mapsto \mathbb{R}$ parameterize a Gaussian distribution $\theta \mapsto \mathcal{N}\left(\mu(\theta), \sigma^2(\theta)\right)$. Then the diagonal of the Fisher information matrix $\mathcal{I}$ is
    \begin{equation} \label{eq:fisher_diagonal}
        \mathcal{I}_{kk} = \left(\frac{\partial \mu}{\partial \theta_k}\cdot\frac{1}{\sigma}\right)^{2}+
        2\left(\frac{\partial \sigma}{\partial \theta_k}\cdot\frac{1}{\sigma}\right)^{2}.
    \end{equation}
\end{fact}

Note that in SAC we use a factorized multivariate Gaussian distribution with mean vector $(\mu_1, \ldots, \mu_N)$ and standard deviations $(\sigma_1, \ldots, \sigma_N)$. It is straightforward to check that this case
\[
    \mathcal{I}_{kk} = \sum_{\ell=1}^N \left(\frac{\partial \mu_\ell}{\partial \theta_k}\cdot\frac{1}{\sigma_\ell}\right)^{2}+
    2\left(\frac{\partial \sigma_\ell}{\partial \theta_k}\cdot\frac{1}{\sigma_\ell}\right)^{2}.
\]

% EWC uses the Fisher information matrix to approximate the importance of each parameter in the network. Calculating the Fisher for classification problems, which are often considered in the supervised setting, is fairly trivial. However, calculating Fisher in the RL setting requires non-obvious assumptions and careful derivations, which we show here. \piotrm{I'd tune down, unless there is a clear non-trivial point. } \maciejw{I think it may be unclear that we assume that value function returns the mean of a Gaussian distribution, at least we discussed other possibilities before having feedback from Razvan. Other than that, I think all is clear.}

\begin{proof} We fix parameters $\theta_0$, denote $\mu_{0} = \mu(\theta_0),\sigma_{0} = \sigma(\theta_0)$ and $\mathcal{N}_{0}:=\mathcal{N}_{0}(\mu_{0},\sigma_{0}^{2})$. We will calculate the the diagonal Fisher matrix coefficient at $\theta_0$ using the fact that it is the curvature of the  Kullback–Leibler divergence:
    \begin{equation} \label{eq:dkl_curvature}
        \mathcal{I}_{kk} = \frac{\partial^2 }{\partial \theta_k^2} F(\theta), \quad F(\theta) := f(\mu(\theta_{i}),\sigma(\theta_{i})),
    \end{equation}
    where
    \[
        f(\mu,\sigma):={\displaystyle D_{\text{KL}}\big({\mathcal{N}}_{0}\parallel{\mathcal{N}}(\mu, \sigma^2)\big)=\frac{1}{2}\left\{ \left({\frac{\sigma_{0}}{\sigma}}\right)^{2}+{\frac{(\mu-\mu_{0})^{2}}{\sigma^{2}}}-1+2\ln\frac{\sigma}{\sigma_{0}}\right\} }.
    \]
    We start with
    \begin{equation}
        \frac{\partial f}{\partial\mu}=\frac{(\mu-\mu_{0})}{\sigma^{2}},\quad\text{thus}\quad\frac{\partial f}{\partial\mu}(\mu_{0},\sigma_{0})=0.\label{eq:dmu}
    \end{equation}
    Further
    \begin{equation}
        \frac{\partial^{2}f}{\partial\mu^{2}}=\frac{1}{\sigma^{2}},\quad\text{thus}\quad\frac{\partial f}{\partial\mu}(\mu_{0},\sigma_{0})=\frac{1}{\sigma_{0}^{2}}.\label{eq:d2mu}
    \end{equation}
    For $\sigma$ we have
    % \begin{equation}
    %     \frac{\partial f}{\partial\sigma}=\frac{1}{2}\left\{ -2\frac{\sigma_{0}^{2}}{\sigma^{3}}-2\frac{(\mu-\mu_{0})^{2}}{\sigma^{3}}+\frac{2}{\sigma}\right\} =-\frac{\sigma_{0}^{2}}{\sigma^{3}}-\frac{(\mu-\mu_{0})^{2}}{\sigma^{3}}+\frac{1}{\sigma},\quad\text{thus}\quad\frac{\partial f}{\partial\sigma}(\mu_{0},\sigma_{0})=-\frac{\sigma_{0}^{2}}{\sigma_{0}^{3}}-\frac{0}{\sigma^{3}}+\frac{1}{\sigma_{0}}=0.
    % \end{equation}
    \begin{align}\label{eq:dsigma}
        &\frac{\partial f}{\partial\sigma}=\frac{1}{2}\left\{ -2\frac{\sigma_{0}^{2}}{\sigma^{3}}-2\frac{(\mu-\mu_{0})^{2}}{\sigma^{3}}+\frac{2}{\sigma}\right\} =-\frac{\sigma_{0}^{2}}{\sigma^{3}}-\frac{(\mu-\mu_{0})^{2}}{\sigma^{3}}+\frac{1}{\sigma},\\
        &\frac{\partial f}{\partial\sigma}(\mu_{0},\sigma_{0})=-\frac{\sigma_{0}^{2}}{\sigma_{0}^{3}}-\frac{0}{\sigma^{3}}+\frac{1}{\sigma_{0}}=0.\nonumber
    \end{align}
    Further
    \begin{equation}
        \frac{\partial^{2}f}{\partial\sigma^{2}}=3\frac{\sigma_{0}^{2}}{\sigma^{4}}+3\frac{(\mu-\mu_{0})^{2}}{\sigma^{4}}-\frac{1}{\sigma^{2}},\quad\text{thus}\quad\frac{\partial^{2}f}{\partial\sigma^{2}}(\mu_{0},\sigma_{0})=3\frac{\sigma_{0}^{2}}{\sigma_{0}^{4}}+3\frac{0}{\sigma^{4}}-\frac{1}{\sigma_{0}^{2}}=\frac{2}{\sigma_{0}^{2}}.\label{eq:d2sigma}
    \end{equation}
    Now, we come back to \eqref{eq:dkl_curvature}. Applying the chain rule
    \[
        \frac{\partial^2 F}{\partial \theta_k^2} =
        \frac{\partial^2 \mu}{\partial \theta_k^2}\cdot\frac{\partial f}{\partial\mu}+
        \left(\frac{\partial \mu}{\partial \theta_k}\right)^2\cdot\frac{\partial^2 f}{\partial\mu^2}+
        \frac{\partial^2 \sigma}{\partial \theta_k^2}\cdot\frac{\partial f}{\partial\sigma}+
        \left(\frac{\partial \sigma}{\partial \theta_k}\right)\cdot\frac{\partial^2 f}{\partial\sigma^2}.
    \]
    When evaluating at $\theta = \theta_0$ the first and third terms vanish, by \eqref{eq:dmu} and \eqref{eq:dsigma}. Thus
    \[
        \frac{\partial^2 F}{\partial \theta_k^2}(\theta_0) =
        \left(\frac{\partial \mu}{\partial \theta_k}(\theta_0)\right)^2\cdot\frac{\partial^2 f}{\partial\mu^2}(\theta_0)+
        \left(\frac{\partial \sigma}{\partial \theta_k}(\theta_0)\right)^2\cdot\frac{\partial^2 f}{\partial\sigma^2}(\theta_0).
    \]
    Now we use (\ref{eq:dsigma}) and (\ref{eq:d2sigma}) we arrive at \eqref{eq:fisher_diagonal}.
    % \[
    %     \frac{\partial^2 F}{\partial \theta_k^2}(\theta_0) =
    %     \left(\frac{\partial \mu}{\partial \theta_k}(\theta_0)\cdot\frac{1}{\sigma(\theta^{0})}\right)^{2}+
    %     2\left(\frac{\partial \sigma}{\partial \theta_k}(\theta_0)\cdot\frac{1}{\sigma(\theta^{0})}\right)^{2}.
    % \]

\end{proof}

\subsection{VCL}% KL weight coefficient and prior for the first task}
\label{sec:vcl_problems_appendix}

\begin{figure}
    \centering
    \includegraphics[width=\textwidth]{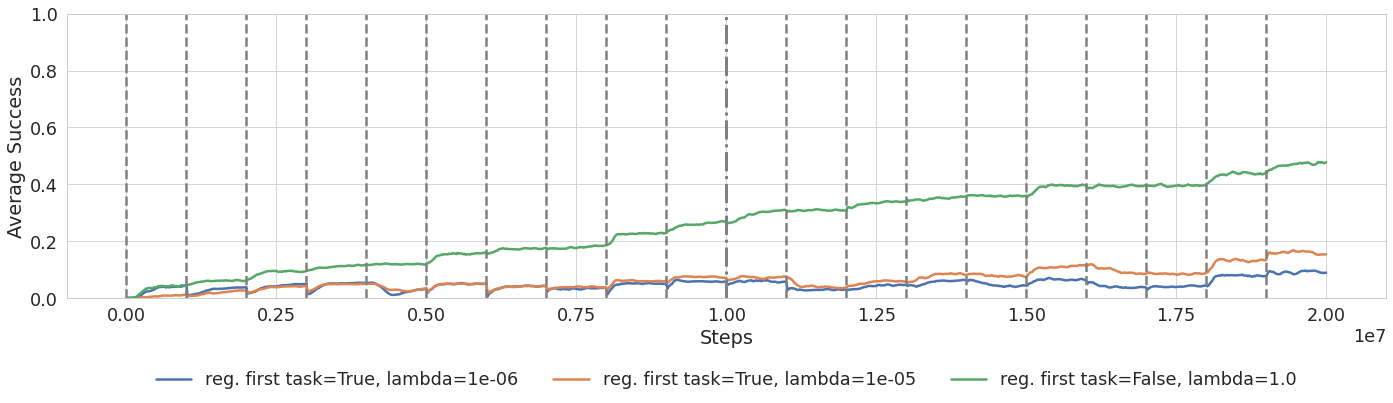}
    \caption{Average performance for VCL.}
    \label{fig:vcl_problems_appendix}
\end{figure}

We suggest introducing two changes to the original VCL setup \cite{DBLP:conf/iclr/NguyenLBT18}: treating $\lambda$ as hyperparameter and not using prior when training the first task. Our conclusion follows from the following three experimental evaluations:

% VCL was derived from the Bayesian perspective of maximizing likelihood along with incorporating strong priors. However, reinforcement learning is significantly different than usual MLE settings \maciejw{Is there a paper here which we can cite?}. Because of that and based on empirical evaluation we decided to introduce small changes to VCL. In this section we compare VCL as implemented in the original paper to VCL with our modified version adapted for the RL setting. In particular, we focus on two points which we changed: (1) performing a hyperparameter search for $\lambda$ instead of a fixed $\lambda$, and (2) not using the prior $\mathcal{N}(0, 1)$ when training on the first task.

% Thus, we consider three options in our experimental evaluation:
\begin{enumerate}
    \item Fixed $\lambda = 1/N$, where $N=10^6$ is the number of samples in each task.  The prior is applied during the first task.
    \item We tested the following hyperparameter values: $\lambda \in  \{ 10^{-7}, 10^{-6}, 10^{-5}, 10^{-4}, 10^{-3}, 10^{-2}, 10^{-1}, 10^0 \}$; selected value is $10^{-5}$.  The prior is applied during the first task.
          % $\lambda = 10^{-5}$ obtained via hyperparameter search from $\lambda \in  \{ 10^{-7}, 10^{-6}, 10^{-5}, 10^{-4}, 10^{-3}, 10^{-2}, 10^{-1}, 10^0 \}$\piotrm{change the order}.
    \item We tested the following hyperparameter values: $\lambda \in  \{ 10^{-7}, 10^{-6}, 10^{-5}, 10^{-4}, 10^{-3}, 10^{-2}, 10^{-1}, 10^0 \}$; selected value is $1$.  The prior is \emph{not} applied during the first task.
\end{enumerate}

The first setup is taken from \cite{DBLP:conf/iclr/NguyenLBT18}. As can be seen in Figure \ref{fig:vcl_problems_appendix} it underperformed, failing to keep good performance on previous tasks. Increasing $\lambda$ to $10^{-5}$ improves the results slightly. We observed that higher values of $\lambda$ fail due to strong regularization with respect to the initial prior. Abandoning this prior has led us to the most successful setup. Interestingly, the much higher value of $\lambda=1.0$ is optimal here. %\kucil{I would tone it down a little: ... gives good performance.}

\subsection{Reservoir Sampling}
\label{sec:rsampling_appendix}

\begin{figure}
    \centering
    \includegraphics[width=\textwidth]{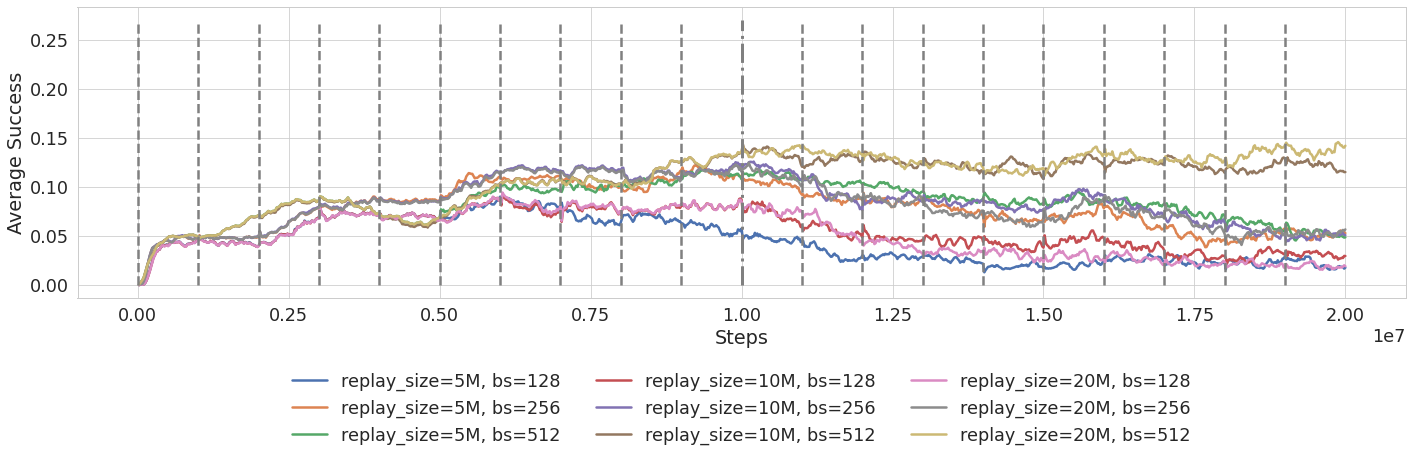}
    \caption{The average performance of Reservoir Sampling. Note the $y$-axis scale change; $bs$ denotes the batch size.}
    \label{fig:reservoir_hps_appendix}
\end{figure}

We present more experiments for the Reservoir Sampling method. We vary the size of the replay $\in\{5\text{M}, 10\text{M}, 20\text{M}\}$. Recall that $20$M is enough to store the whole experience (which corresponds to the Perfect Memory experiments in the main section). For smaller sizes, we use reservoir sampling as described in Section \ref{sec:replay_methods_appendix}. We also test for different batch size values $\{64, 128, 256, 512\}$. We observe that higher values result in better performance, possibly due to better data balance in a single batch. This, however, comes at the cost of training speed.

Figure \ref{fig:reservoir_hps_appendix} shows results for different hyperparameter settings. Runs with the standard batch size $128$ used by the rest of the methods fail to exceed $0.05$ final average success obtained by Fine-tuning. Increasing the batch size improves the results significantly, and the two best runs using batch size $512$ achieve final average performance between $0.1$ and $0.15$. However, we note that even the best performing hyperparameters for Reservoir Sampling do not achieve as good results as the regularization and parameter isolation methods. We attribute this phenomenon to the critic regularization problem discussed in Section \ref{sec:critic_regularizaiton_appendix}.
% In the main section of the paper, we focused on a particular replay setting called perfect memory, where the agent is able to store all examples from the past. We also allowed the method to sample a larger number of samples from the buffer than in our standard setting (512 vs. 128), which improved results at cost of increased training speed. %Here, we show additional experiments without these assumptions. As explained in Section \ref{sec:replay_methods_appendix}, we use reservoir sampling to choose which samples should be in the buffer, instead of the standard FIFO approach of regular SAC.

\subsection{A-GEM not working}
\label{sec:agem_appendix}

\begin{figure}
    \begin{center}
        \includegraphics[width=\textwidth]{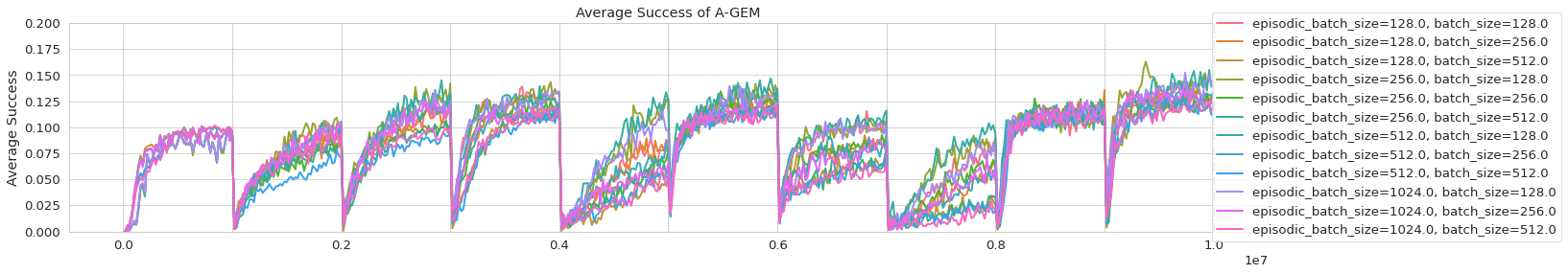}
        \caption{\label{fig:agem_hparams} Average performance of A-GEM on CW10. Note the $y$-axis scale change. }
    \end{center}
\end{figure}

Despite our best efforts, A-GEM has poor performance on our benchmark. In order to better investigate this case, we conduct additional analysis. In Figure~\ref{fig:agem_hparams} we show average performance on CW10 for various hyperparameter settings. We set the size of episodic task memory to 100k samples per task. We search for batch size (number of samples used to compute $\ell(\theta, \mathcal{B}_{new})$) in $\{ 128, 256, 512 \}$ and for the episodic batch size (number of samples used to compute $\ell(\theta, \mathcal{B}_{old})$) in $\{128, 256, 512, 1024 \}$, see \eqref{eq:agem_projection}. The results show that none of those settings perform significantly better than the others. Similar to Reservoir Sampling, we attribute those problems to the critic regularization discussed in Section \ref{sec:critic_regularizaiton_appendix}. Another conjecture is that due to the high variance of RL gradients (higher than in the supervfised learning setting), approximation \eqref{eq:agem_approx} is too brittle to work well.

% We also point to the fact that gradients in reinforcement learning are known to have high variance, which is not as problematic in supervised learning. This may in turn suggest that the approximation \eqref{eq:agem_approx} of the true constraint does not hold well enough in our setting. \maciejw{added some story. I'm not yet fully convinced about the gradient variance argument.}

\subsection{PackNet extensive clipping} \label{sec:pack_net_appendix_problems}
Initially, PackNet was unstable, degenerating (NaNs and infinities) near the task boundaries. We amended this problem by applying global gradient norm clipping \num{2e-5}. We tune this value so that the learning does not explode and yet the performance is mostly preserved. PackNet instability requires further investigation: we conjecture that frequent changes of the set of optimized parameters might amplify the inherent instability of temporal-difference learning.

% Among the weights managed by PackNet, every task has an equal percentage reserved. All biases and normalization parameters (in LayerNorm) are not managed by PackNet; they remain frozen after learning of the first task. Also, last layer is not managed by PackNet. Every task has its own parameters there.

% Hyperparameter that we checked is number of retraining steps after each task (using only data from buffer). We searched among $\{50000, 100000, 200000\}$ and selected 100000.

% As a technical detail, we use global gradient norm clipping to the maximal value of $0.0001$. Without this clipping, learning explodes after several task changes.

\section{Transfer matrix}\label{sec:forward_transfer_matrices_appendix}
In Table \ref{tab:app:transfer_matrix} we show the transfer matrix corresponding to Figure \ref{fig:transfer_matrix_haatmmap} (Section \ref{sec:task_sets}). Recall that the transfer matrix is aimed to measure the relationship between each two tasks 
(for transfer on  \cwtwenty{} see Table \ref{tab:transfer_table}). 
%
%Notice that there are only a few negative forward transfer cases, and those are of a rather small magnitude (perhaps  unsurprisingly, as the tasks are related). There are also visible patterns in the matrix. For instance, some tasks such as \textsc{peg-unplug-side-v1} or \textsc{push-back-v1} benefit from a relatively large forward transfer, (almost) irrespective of the first task. Furthermore, 
Most entries in the Table \ref{tab:app:transfer_matrix} are positive, however, some tasks seem to interfere with one another (most notably \textsc{peg-unplug-v1} (9) followed by \textsc{push-v1} (6)). Some of the factors influencing the transfer are: similarity in visual representation (e.g. objects on the scene), reward structure (reach, grasp, place), or direction in which the robotic arm moves. 
The average forward transfer given the second task (columns) is more variable than the corresponding quantity for the first task (rows), with the standard deviation equal to $0.21$ and $0.07$, respectively. %\kucil{let someone review this}

%\kucil{notice: place-shelf (7) and peg-unplug (9) as well also have rewards for reach, grasp and place, but perform so-so in the matrix.}

\begin{table}[h]
    \centering
\small
\rotatebox{90}{
\begin{tabular}{p{1.1cm}p{.9cm}p{.9cm}p{.9cm}p{.9cm}p{.9cm}p{.9cm}p{.9cm}p{.9cm}p{.9cm}p{.9cm}|p{.6cm}}
\toprule
Second task &                    0 &                     1 &                  2 &                   3 &                    4 &                  5 &                     6 &                  7 &                  8 &                    9 & mean \\
First task &                      &                       &                    &                     &                      &                    &                       &                    &                    &                     & \\
\midrule
0          &    0.42 [0.32, 0.51] &    0.13 [-0.01, 0.24] &  0.64 [0.56, 0.70] &   0.58 [0.35, 0.76] &    0.24 [0.12, 0.35] &  0.48 [0.39, 0.56] &    0.03 [-0.09, 0.14] &  0.58 [0.54, 0.62] &  0.32 [0.26, 0.38] &    0.47 [0.05, 0.81]&0.39 \\
1          &    0.13 [0.01, 0.25] &   -0.05 [-0.18, 0.05] &  0.32 [0.11, 0.48] &   0.76 [0.72, 0.79] &   0.01 [-0.09, 0.11] &  0.46 [0.38, 0.53] &   -0.04 [-0.12, 0.04] &  0.28 [0.21, 0.35] &  0.36 [0.30, 0.42] &    0.78 [0.69, 0.85]&0.30 \\
2          &   0.04 [-0.08, 0.15] &   -0.04 [-0.10, 0.03] &  0.50 [0.35, 0.63] &   0.75 [0.72, 0.78] &   0.01 [-0.08, 0.10] &  0.26 [0.19, 0.33] &  -0.17 [-0.30, -0.06] &  0.32 [0.22, 0.41] &  0.27 [0.17, 0.35] &    0.79 [0.71, 0.85]&0.27 \\
3          &   0.03 [-0.22, 0.24] &  -0.15 [-0.30, -0.01] &  0.40 [0.14, 0.61] &   0.47 [0.23, 0.67] &   0.09 [-0.00, 0.18] &  0.42 [0.34, 0.49] &    0.01 [-0.10, 0.11] &  0.39 [0.29, 0.48] &  0.32 [0.24, 0.40] &   0.33 [-0.08, 0.66]&0.23 \\
4          &   0.19 [-0.02, 0.36] &     0.21 [0.14, 0.28] &  0.45 [0.20, 0.63] &   0.47 [0.22, 0.70] &    0.42 [0.34, 0.50] &  0.41 [0.32, 0.50] &     0.19 [0.11, 0.26] &  0.53 [0.46, 0.59] &  0.36 [0.28, 0.43] &   0.38 [-0.05, 0.71]&0.36 \\
5          &   0.05 [-0.08, 0.16] &   -0.00 [-0.14, 0.11] &  0.55 [0.43, 0.65] &   0.64 [0.49, 0.74] &   0.05 [-0.04, 0.13] &  0.51 [0.45, 0.56] &  -0.14 [-0.25, -0.05] &  0.20 [0.06, 0.33] &  0.32 [0.25, 0.38] &  -0.01 [-0.61, 0.50]&0.22 \\
6          &    0.20 [0.08, 0.31] &   -0.06 [-0.16, 0.03] &  0.47 [0.35, 0.57] &   0.78 [0.75, 0.81] &   0.09 [-0.03, 0.19] &  0.55 [0.50, 0.60] &   -0.01 [-0.08, 0.06] &  0.30 [0.18, 0.41] &  0.42 [0.35, 0.48] &    0.81 [0.77, 0.85]&0.36 \\
7          &    0.29 [0.09, 0.44] &   -0.03 [-0.14, 0.07] &  0.33 [0.07, 0.53] &  0.20 [-0.10, 0.47] &    0.14 [0.04, 0.24] &  0.42 [0.35, 0.48] &   -0.06 [-0.17, 0.05] &  0.40 [0.31, 0.49] &  0.33 [0.24, 0.41] &   0.10 [-0.48, 0.57]&0.21 \\
8          &  -0.10 [-0.33, 0.10] &     0.06 [0.00, 0.12] &  0.24 [0.04, 0.41] &   0.69 [0.63, 0.74] &  -0.01 [-0.11, 0.08] &  0.22 [0.15, 0.28] &   -0.02 [-0.08, 0.04] &  0.42 [0.34, 0.49] &  0.36 [0.29, 0.43] &    0.80 [0.75, 0.84]&0.27 \\
9          &   0.10 [-0.09, 0.26] &   -0.05 [-0.15, 0.04] &  0.30 [0.06, 0.48] &   0.53 [0.34, 0.68] &  -0.02 [-0.11, 0.08] &  0.18 [0.08, 0.26] &  -0.28 [-0.40, -0.16] &  0.21 [0.12, 0.30] &  0.31 [0.25, 0.37] &    0.74 [0.63, 0.83]&0.20 \\
\midrule
mean& 0.14&	0.00&	0.42	&0.59&	0.10&	0.39	&-0.05&	0.36	&0.34	&0.52& 0.28\\
\bottomrule
\end{tabular}}
    \caption{Transfer matrix. Normalized forward transfers for the second task. For task names see Section \ref{sec:sequences_appendix}.}
    \label{tab:app:transfer_matrix}
\end{table}

% \piotrm{Put some story}

\subsection{Diagonal transfers}\label{sec:diagonal_transfers_appendix}
The transfers on the diagonal (i.e., between the same tasks) are smaller than expected. We enumerate possible reasons: using multi-head architecture, buffer resetting, and uniform sampling exploration policy for the second task. These design elements are generally required when switching tasks, but not on the diagonal, where we resume the same task. This makes the diagonal of the transfer matrix suitable for an in-depth ablation analysis.

\begin{figure}
    \centering
    \includegraphics[width=\textwidth]{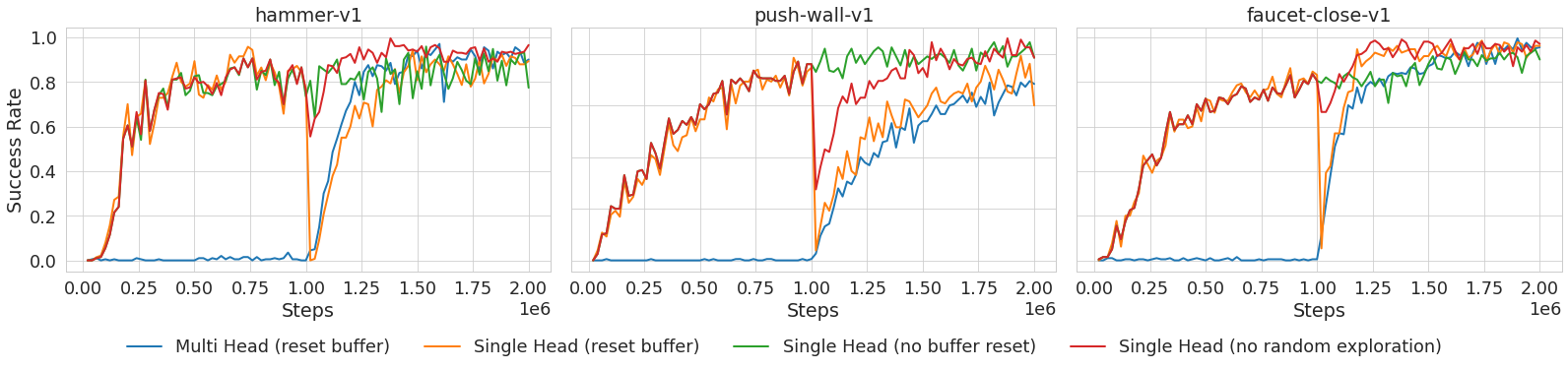}
    \caption{Success rates for example tasks on the diagonal of the transfer matrix. The blue line is the standard setup. Note, that its poor performance in the first part of each plot is expected, as we evaluate the head which has not yet been trained. Performance improves when we keep the samples in the buffer when switching the task or when we disable the exploration through uniform policy.}
    \label{fig:diagonal_appendix}
\end{figure}

% In Table \ref{tab:diagonal_transfer} and example curves in in Figure \ref{fig:diagonal_appendix} we present results for four settings.
The experiments for the whole diagonal are summarized in Table \ref{tab:diagonal_transfer} and example curves are shown in Figure \ref{fig:diagonal_appendix}. We tested four settings. Multi Head (reset buffer) is our standard setting. Changing to a single head network, dubbed as Single Head (reset buffer), brings no significant difference. However, keeping the examples in the buffer in Single Head (no buffer reset) yields a major improvement. Similarly, in Single Head (no random exploration), when we do not perform random exploration at the task change the performance improves. We conjecture, that resetting the buffer and performing random exploration results in distributional shift, which is hard to be handled by SAC. %\piotrm{Verify please. Possilby we can also plug off-policy story.} \maciejw{Sounds good.}

% We found that resetting the buffer has the biggest impact on the performance of the second task (i.e., the one we transfer to). Using the multiple heads and random exploration is of smaller relevance. \maciejw{It is not exactly true that random exploration is of smaller relevance and I think we should explicitly say why we do not use it.}

\begin{table}\scriptsize
    \centering
    \begin{tabular}{lrrrr}
        \toprule
        Task                 & MH (reset buffer) & SH (reset buffer) & SH (no buffer reset) & SH (no random exp) \\
        \midrule
        \texttt{hammer-v1}            & 0.42              & 0.26              & 0.56                 & 0.74               \\
        \texttt{push-wall-v1}         & -0.05             & 0.06              & 0.55                 & 0.40               \\
        \texttt{faucet-close-v1}      & 0.50              & 0.64              & 0.60                 & 0.78               \\
        \texttt{push-back-v1}         & 0.47              & 0.41              & 0.67                 & 0.68               \\
        \texttt{stick-pull-v1}        & 0.42              & 0.37              & 0.60                 & 0.63               \\
        \texttt{handle-press-side-v1} & 0.51              & 0.53              & 0.53                 & 0.68               \\
        \texttt{push-v1}              & -0.01             & -0.03             & 0.64                 & 0.35               \\
        \texttt{shelf-place-v1}       & 0.40              & 0.23              & 0.29                 & 0.40               \\
        \texttt{window-close-v1}      & 0.36              & 0.50              & 0.45                 & 0.52               \\
        \texttt{peg-unplug-side-v1}   & 0.74              & 0.83              & 0.97                 & 0.96               \\
        \midrule
        mean              & 0.38              & 0.38              & 0.58                 & 0.61               \\
        \bottomrule
    \end{tabular}
    \caption{Transfers on the diagonal. MH and SH stand for multi-head and single-head, respectively.}\label{tab:diagonal_transfer}
\end{table}

% \begin{itemize}
%     % \item The transfers on the diagonal are not very high, sometimes they are even slightly negative (push-wall-v1 (id 1) = -0.05, push-v1 (id 6) = -0.01). Why is that?
%     \item We examine whether switching form multi-head (blue lines) to single-head (orange line) helps. It sometimes helps, but there is still a significant "collapse" at the point when we change task.
%     % \item We then test what happens if we do not reset the buffer samples (green line). There is no collapse, forward transfer is much higher. This is cheating in some sense - this approach is incompatible with the CL mindset of not having samples from the previous task and would not work outside the diagonal (reservoir sampling fails in our regular experiments).
%     \item In SAC if we do not have any samples in the buffer, we use uniform sampling, let's use our standard policy (red line) instead. This works better then orange line but sometimes worse than green line. This would hinder exploration outside the diagonal.
%     \item Third plot is especially interesting - seems like dropping old samples actually helps here (red line higher than green line).
% \end{itemize}
\newpage
\section{CW20 results}\label{sec:cwtwenty_graph_and_analysis_appendix}
In Table \ref{tab:collected_resuls_appendix}, we recall for convenience the summary of our results for the experiments on the \cwtwenty{} sequence (additional backward transfer results are described in Section \ref{sec:forgetting_appendix}). We also recall that the methods were tuned with the objective of maximizing the final average performance. Our main findings are that most methods are efficient with mitigating forgetting but have an unsatisfying forward transfer. The best method, PackNet, has forward transfer close to Fine-tuning, which is a strong baseline for this metric. However, both fall below the reference value obtained from the analysis of the transfer matrix $\text{RT}=0.46$.

Table \ref{tab:collected_resuls_appendix} additionally highlights an important fact that CL metrics are interrelated. For instance, high forward transfer and low performance (a case for Fine-tuning and A-GEM) have to imply high forgetting.

\begin{table}[ht!]
    \begin{center}
        \tabcolsep=0.11cm
        \renewcommand{\arraystretch}{0.9}
        \begin{tabular}{lllll}
            \toprule
            method                  & performance                         & forgetting                  & f. transfer                         & b. transfer               \\
            \midrule
            \textbf{Fine-tuning}        & 0.05 {\tiny [0.05, 0.06]} & 0.73 {\tiny [0.72, 0.75]}   & \textbf{0.20 {\tiny [0.17, 0.23]}} & 0.00 {\tiny[0.00, 0.00]} \\
            \textbf{L2}          & 0.43 {\tiny [0.39, 0.47]} & 0.02 {\tiny [0.00, 0.03]}   & -0.71 {\tiny [-0.87, -0.57]}      & 0.02 {\tiny[0.02, 0.03]} \\
            \textbf{EWC}         & 0.60 {\tiny [0.57, 0.64]} & 0.02 {\tiny [-0.00, 0.05]}  & -0.17 {\tiny [-0.24, -0.11]}       & 0.04 {\tiny[0.03, 0.04]} \\
            \textbf{MAS}         & 0.51 {\tiny [0.49, 0.53]} & 0.00 {\tiny [-0.01, 0.02]}  & -0.52 {\tiny [-0.59, -0.47]}      & 0.04 {\tiny[0.04, 0.05]} \\
            \textbf{VCL}         & 0.48 {\tiny [0.46, 0.50]} & 0.01 {\tiny [-0.01, 0.02]}  & -0.49 {\tiny [-0.57, -0.42]}       & 0.04 {\tiny[0.03, 0.04]} \\
            \textbf{PackNet}     & \textbf{0.80 {\tiny [0.79, 0.82]}} & 0.00 {\tiny [-0.01, 0.01]} & \textbf{0.19 {\tiny [0.15, 0.23]}} & 0.03 {\tiny[0.03, 0.04]} \\
            \textbf{Perfect Memory}   & 0.12 {\tiny [0.09, 0.15]} & 0.07 {\tiny [0.05, 0.10]}   & -1.34 {\tiny [-1.42, -1.27]}       & 0.01 {\tiny[0.01, 0.01]} \\
            \textbf{A-GEM}        & 0.07 {\tiny [0.06, 0.08]} & 0.71 {\tiny [0.70, 0.73]}   & 0.13 {\tiny [0.10, 0.16]}          & 0.00 {\tiny[0.00, 0.00]} \\
            \midrule
            \textbf{MT}         & 0.51 {\tiny [0.48, 0.53]} & ---                 & ---                                 & ---                       \\
            \textbf{MT (PopArt)}  & 0.65 {\tiny [0.63, 0.67]} & ---                 & ---                                 & ---                       \\
            \midrule
            \textbf{RT}             & ---                                 & ---                         & \textbf{0.46}                       & ---                       \\
            \bottomrule
        \end{tabular}
        \caption{Results on \cwtwenty{}, for CL methods and multi-task training.}
        \label{tab:collected_resuls_appendix}
    \end{center}
\end{table}

We recall that CW10 and CW20 sequences are defined in Section \ref{sec:sequences_appendix}. In the next sections, we discuss the results in more detail. We also provide the following visualizations:
\begin{itemize}
    \item Figure \ref{sec:average_success_big} - performance curves averaged over tasks.
    \item Figure \ref{fig:current_task_big} - performance curves for the active task.
    \item Figure \ref{sec:task_success_big} - performance curves for all tasks. Useful for qualitative studies of transfer and forgetting.
    \item Figure \ref{fig:foraward_transfer_per_task} - visualization of forward transfer for each task.
    \item Figure \ref{fig:forgetting_big_plot} - forgetting curves averaged over tasks.
\end{itemize}

% \piotrm{Possibly make list of figures.}

% \kucil{pull with stick, place shelf, hammer, unplug peg have the "richest" reward structure (reach, grasp, place). from hammer (and from push back) transfer is positive (transfer matrix)}

\subsection{Forgetting and backward transfer} \label{sec:forgetting_appendix}
\begin{figure}
    \begin{center}
        \includegraphics[width=\textwidth]{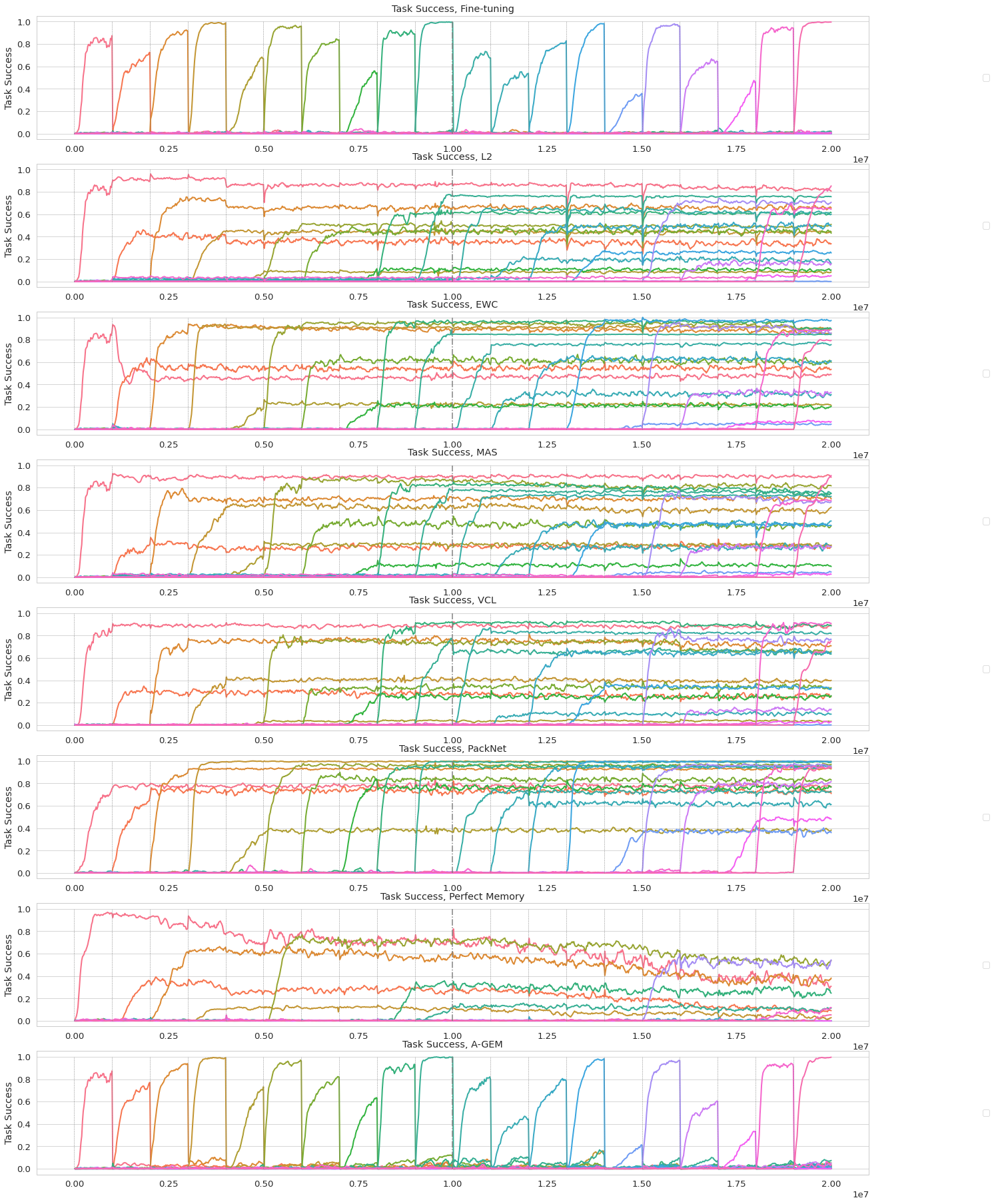}
        \caption{Success rates for all tasks.}\label{sec:task_success_big}
    \end{center}
\end{figure}

\begin{figure}
    \begin{center}
        \includegraphics[width=\textwidth]{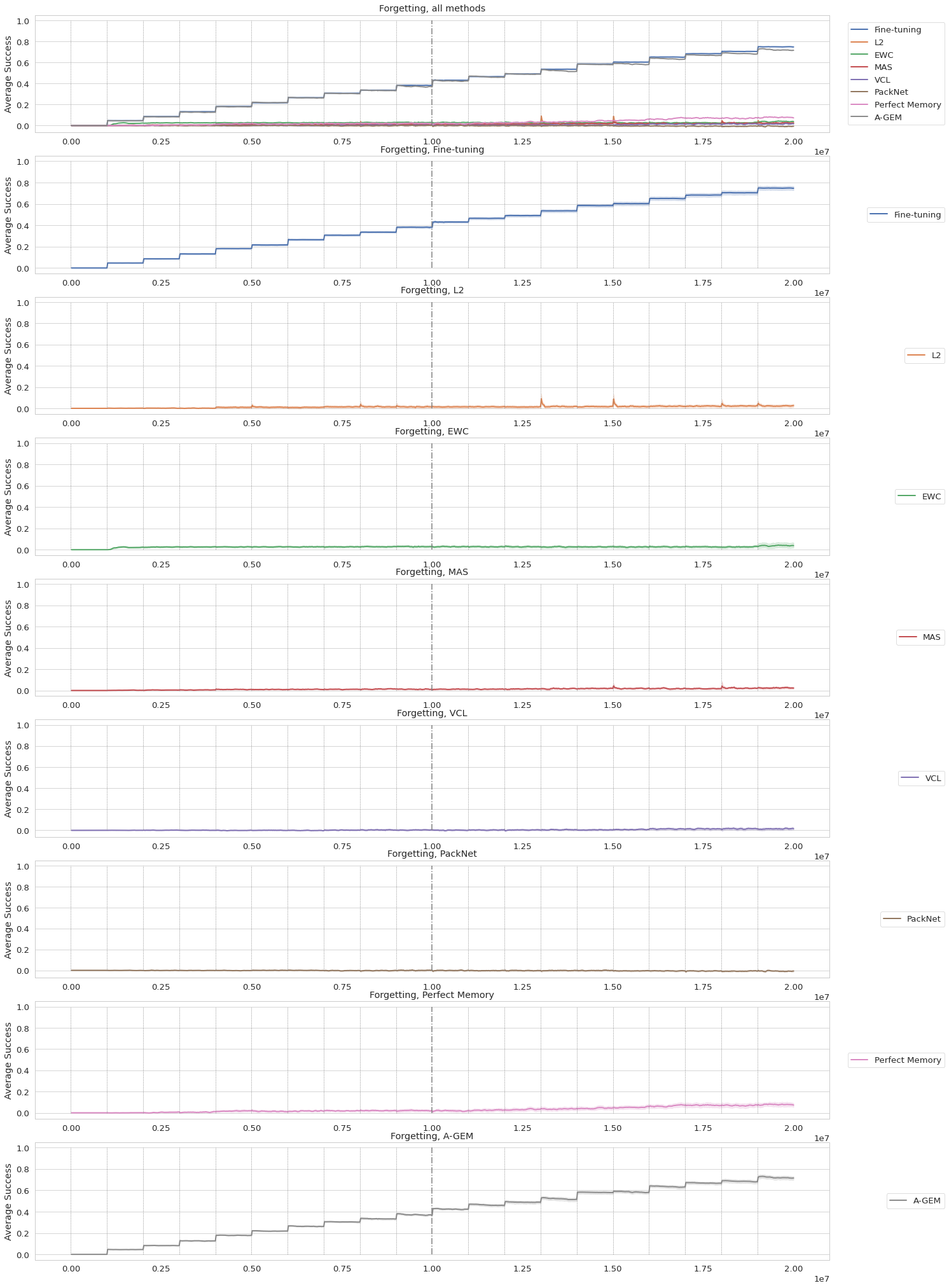}
        \caption{Average (over task) forgetting. For the task $i$ we set forgetting at time $t$ to be $\text{F}_i(t):= p_i(i\cdot \Delta) - p_i(t)$ if $t\geq i\cdot \Delta$ and $0$ otherwise. See also \eqref{eq:forgetting}. } \label{fig:forgetting_big_plot}
    \end{center}
\end{figure}

%Recall Table \ref{tab:collected_resuls_appendix},  in 
Table \ref{tab:forgetting_table} expands upon Table \ref{tab:collected_resuls_appendix}   by presenting more detailed forgetting results, see also Figure \ref{fig:forgetting_big_plot} and Figure \ref{sec:task_success_big}. The latter is convenient to observe the evaluation dynamics of each task.

Fine-tuning and AGEM exhibit rapid catastrophic forgetting after task switch (see Figure \ref{sec:task_success_big}). It is expected for the former, but quite surprising for the latter. We conjecture possible reasons in Section \ref{sec:agem_appendix}. Some mild forgetting can also be observed for Perfect Memory (it is not directly observed in the metric due to the poor training but can be visible on the graphs). The rest of the methods are quite efficient in mitigating forgetting. This also includes the very basic L2 method (which is at the cost of poor transfer, however).

In our experiment, we do not observe backward transfer, see Table \ref{tab:collected_resuls_appendix}. The values were calculated according to $B := \frac{1}{N}\sum_{i=1}^N B_i$, where
\[
    B_i := \max\left\{0,  p_i(T) - p_i(i\cdot \Delta)\right\},
\]
see Section \ref{sec:metrics} for notation. Note that some small values are possible due to stochastic evaluations.

\begin{table}[h]
    \centering
    \scriptsize
    \rotatebox{90}{
    \begin{tabular}{lllllllll}
        \toprule
        task & Fine-tuning       & L2                   & EWC                  & MAS                  & VCL                  & PackNet              & Perfect Memory      & A-GEM             \\
        \midrule
        0    & 0.86 [0.82, 0.90] & 0.07 [-0.04, 0.19]   & 0.38 [0.23, 0.53]    & -0.02 [-0.07, 0.04]  & -0.05 [-0.10, 0.01]  & -0.01 [-0.09, 0.06]  & 0.56 [0.48, 0.64]   & 0.86 [0.82, 0.91] \\
        1    & 0.72 [0.67, 0.77] & 0.08 [0.02, 0.14]    & 0.04 [-0.04, 0.13]   & 0.01 [-0.07, 0.08]   & 0.04 [-0.02, 0.09]   & -0.03 [-0.07, 0.00]  & 0.29 [0.19, 0.40]   & 0.78 [0.72, 0.83] \\
        2    & 0.91 [0.88, 0.95] & 0.09 [0.01, 0.18]    & 0.01 [-0.05, 0.08]   & 0.01 [-0.07, 0.09]   & 0.01 [-0.05, 0.07]   & -0.03 [-0.06, -0.01] & 0.07 [0.03, 0.12]   & 0.92 [0.88, 0.96] \\
        3    & 0.99 [0.98, 0.99] & -0.00 [-0.02, 0.02]  & 0.04 [0.01, 0.07]    & 0.00 [-0.07, 0.08]   & -0.04 [-0.08, -0.00] & -0.01 [-0.01, -0.00] & 0.04 [0.00, 0.12]   & 0.98 [0.96, 0.99] \\
        4    & 0.67 [0.55, 0.77] & -0.01 [-0.03, 0.00]  & -0.03 [-0.08, 0.02]  & -0.10 [-0.17, -0.05] & -0.01 [-0.03, 0.01]  & -0.03 [-0.09, 0.02]  & 0.00 [0.00, 0.00]   & 0.76 [0.67, 0.84] \\
        5    & 0.96 [0.95, 0.97] & -0.01 [-0.03, 0.00]  & 0.01 [-0.06, 0.11]   & -0.01 [-0.10, 0.07]  & 0.13 [0.01, 0.25]    & -0.02 [-0.03, 0.00]  & 0.04 [-0.01, 0.08]  & 0.97 [0.95, 0.99] \\
        6    & 0.83 [0.80, 0.86] & 0.02 [-0.02, 0.06]   & 0.02 [-0.02, 0.06]   & 0.04 [0.00, 0.08]    & -0.00 [-0.05, 0.05]  & 0.04 [-0.00, 0.09]   & 0.00 [0.00, 0.00]   & 0.78 [0.75, 0.81] \\
        7    & 0.54 [0.44, 0.64] & 0.01 [-0.02, 0.05]   & 0.01 [-0.04, 0.06]   & 0.01 [-0.02, 0.04]   & -0.02 [-0.06, 0.02]  & 0.06 [-0.00, 0.13]   & 0.00 [0.00, 0.00]   & 0.66 [0.49, 0.80] \\
        8    & 0.87 [0.83, 0.90] & -0.01 [-0.03, 0.02]  & 0.02 [-0.07, 0.12]   & 0.07 [-0.01, 0.17]   & -0.06 [-0.12, 0.01]  & -0.00 [-0.02, 0.01]  & 0.01 [-0.07, 0.08]  & 0.89 [0.84, 0.94] \\
        9    & 0.97 [0.96, 0.98] & 0.01 [-0.01, 0.04]   & -0.00 [-0.01, 0.00]  & 0.03 [-0.04, 0.11]   & 0.11 [0.03, 0.20]    & 0.00 [-0.00, 0.01]   & 0.01 [-0.02, 0.04]  & 0.93 [0.88, 0.98] \\
        10   & 0.67 [0.54, 0.80] & 0.05 [0.00, 0.12]    & -0.08 [-0.12, -0.04] & -0.01 [-0.03, 0.01]  & 0.05 [0.00, 0.12]    & -0.02 [-0.06, 0.03]  & 0.00 [0.00, 0.00]   & 0.66 [0.47, 0.82] \\
        11   & 0.54 [0.49, 0.59] & 0.04 [0.01, 0.07]    & -0.01 [-0.06, 0.04]  & -0.01 [-0.05, 0.03]  & -0.00 [-0.02, 0.02]  & 0.08 [0.01, 0.15]    & 0.00 [0.00, 0.00]   & 0.42 [0.32, 0.50] \\
        12   & 0.83 [0.78, 0.87] & 0.02 [-0.02, 0.06]   & 0.00 [-0.05, 0.07]   & -0.01 [-0.04, 0.02]  & -0.01 [-0.05, 0.03]  & 0.00 [-0.03, 0.04]   & 0.00 [0.00, 0.01]   & 0.70 [0.60, 0.80] \\
        13   & 0.98 [0.97, 0.99] & 0.01 [0.00, 0.02]    & -0.02 [-0.06, 0.01]  & -0.00 [-0.03, 0.02]  & -0.00 [-0.04, 0.04]  & -0.00 [-0.01, 0.01]  & 0.00 [0.00, 0.00]   & 0.95 [0.89, 0.99] \\
        14   & 0.36 [0.23, 0.49] & 0.00 [0.00, 0.00]    & -0.02 [-0.04, 0.00]  & -0.01 [-0.02, 0.00]  & 0.00 [-0.00, 0.00]   & -0.02 [-0.05, 0.01]  & 0.00 [0.00, 0.00]   & 0.36 [0.18, 0.55] \\
        15   & 0.95 [0.94, 0.97] & -0.02 [-0.06, 0.02]  & 0.09 [0.02, 0.18]    & 0.08 [-0.00, 0.18]   & 0.08 [-0.01, 0.18]   & -0.00 [-0.02, 0.02]  & -0.00 [-0.04, 0.03] & 0.97 [0.95, 0.98] \\
        16   & 0.63 [0.57, 0.69] & 0.00 [-0.03, 0.03]   & -0.01 [-0.04, 0.03]  & -0.01 [-0.05, 0.03]  & -0.02 [-0.05, 0.00]  & -0.03 [-0.06, 0.01]  & 0.00 [0.00, 0.00]   & 0.55 [0.49, 0.61] \\
        17   & 0.46 [0.38, 0.55] & -0.02 [-0.05, -0.00] & 0.00 [-0.02, 0.03]   & 0.00 [-0.01, 0.01]   & -0.01 [-0.03, 0.00]  & -0.05 [-0.09, -0.00] & 0.00 [0.00, 0.00]   & 0.43 [0.30, 0.56] \\
        18   & 0.93 [0.90, 0.96] & 0.02 [-0.03, 0.08]   & -0.03 [-0.06, -0.01] & -0.01 [-0.04, 0.02]  & -0.03 [-0.07, 0.01]  & 0.01 [-0.01, 0.04]   & -0.04 [-0.09, 0.01] & 0.88 [0.81, 0.94] \\
        19   & 0.00 [0.00, 0.00] & 0.00 [0.00, 0.00]    & 0.00 [0.00, 0.00]    & 0.00 [0.00, 0.00]    & 0.00 [0.00, 0.00]    & 0.00 [0.00, 0.00]    & 0.00 [0.00, 0.00]   & 0.00 [0.00, 0.00] \\
        \midrule
        mean & 0.73 [0.72, 0.75] & 0.02 [0.00, 0.03]    & 0.02 [-0.00, 0.05]   & 0.00 [-0.01, 0.02]   & 0.01 [-0.01, 0.02]   & -0.00 [-0.01, 0.01]  & 0.05 [0.04, 0.06]   & 0.72 [0.71, 0.74] \\
        \bottomrule
    \end{tabular}}
    \caption{Forgetting for each method (columns) and tasks (rows). } \label{tab:forgetting_table}
\end{table}

\subsection{Forward transfer} \label{sec:forward_transfer_appendix}
\begin{figure}
    \begin{center}
        \includegraphics[width=\textwidth]{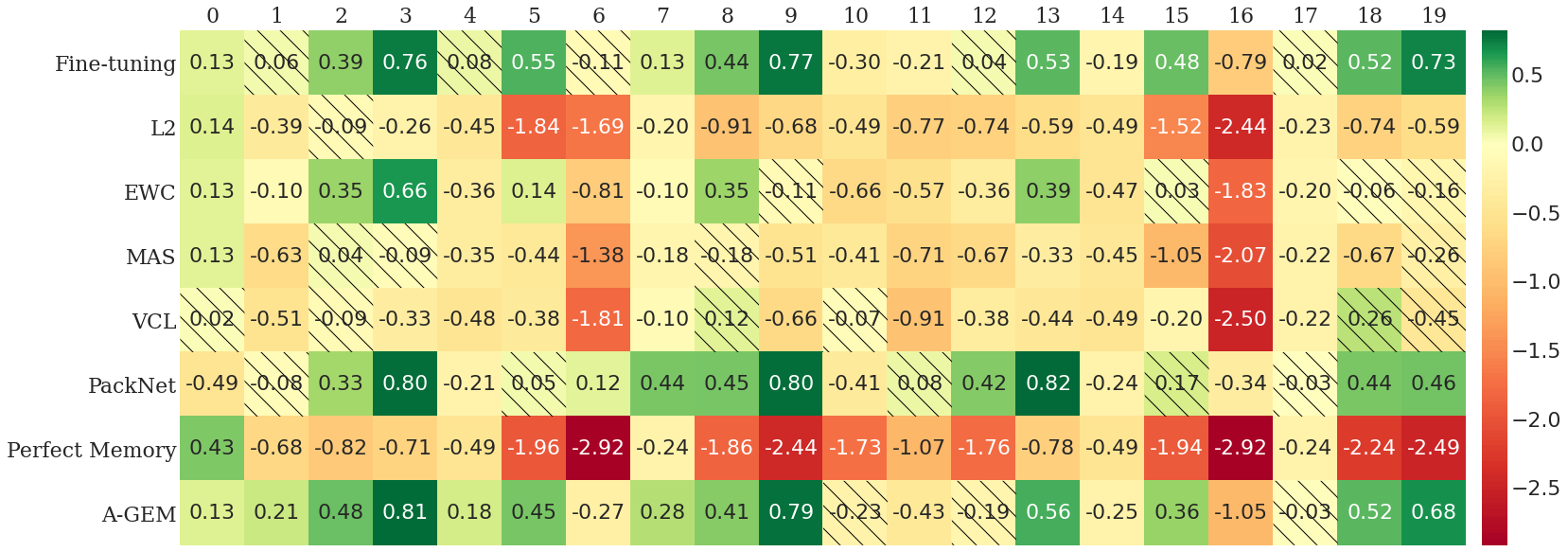}
        \caption{Forward transfer heatmap for CW20. The shaded cells indicate that $0$ belongs to the corresponding $90\%$ confidence interval. }\label{fig:transfer_heatmap}
    \end{center}
\end{figure}

\begin{figure}
    \begin{center}
        \includegraphics[width=\textwidth]{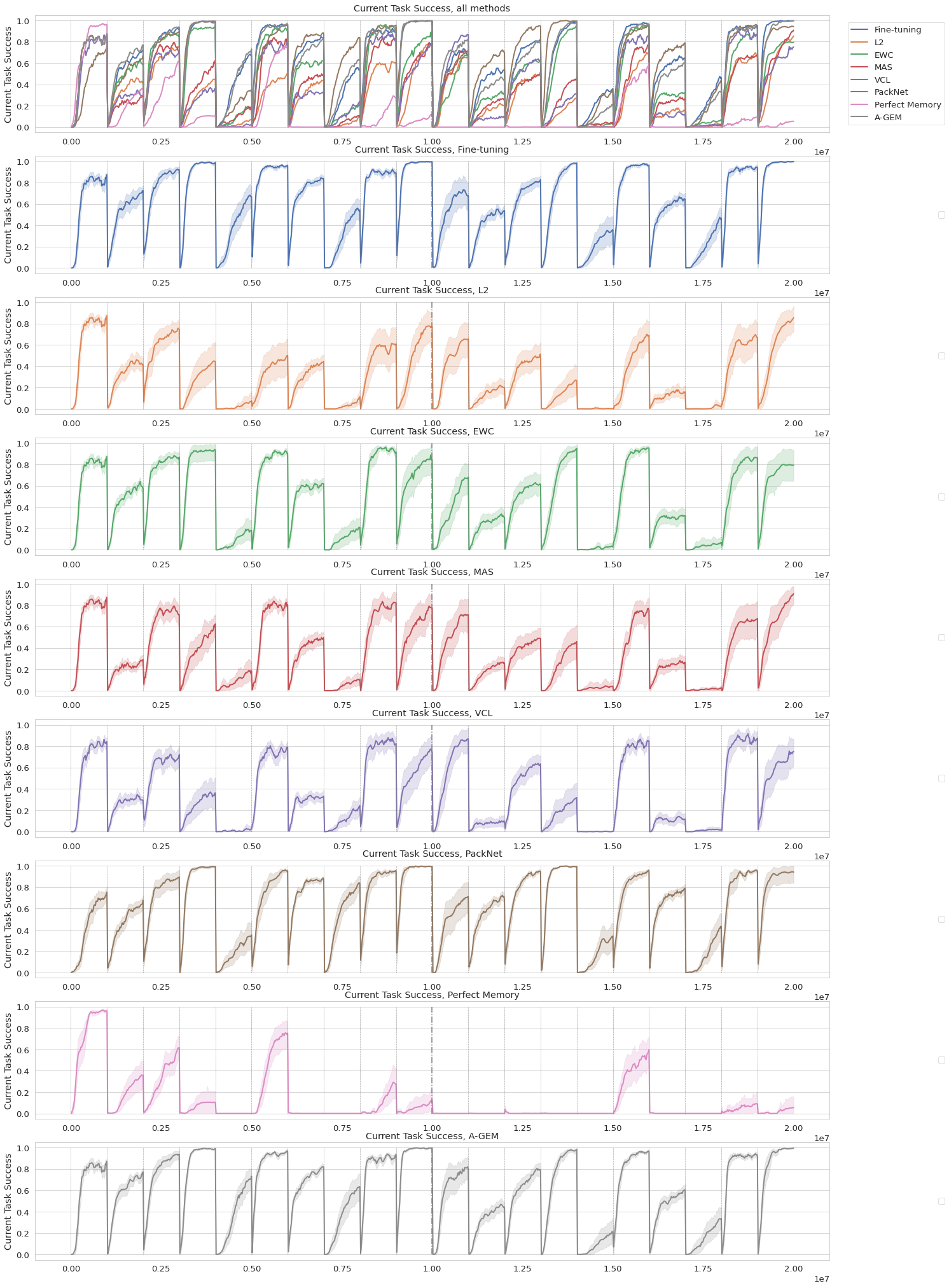}
        \caption{Success rate for the current task (the one being trained).}\label{fig:current_task_big}
    \end{center}
\end{figure}

\begin{figure}
    \begin{center}
        \includegraphics[width=\textwidth]{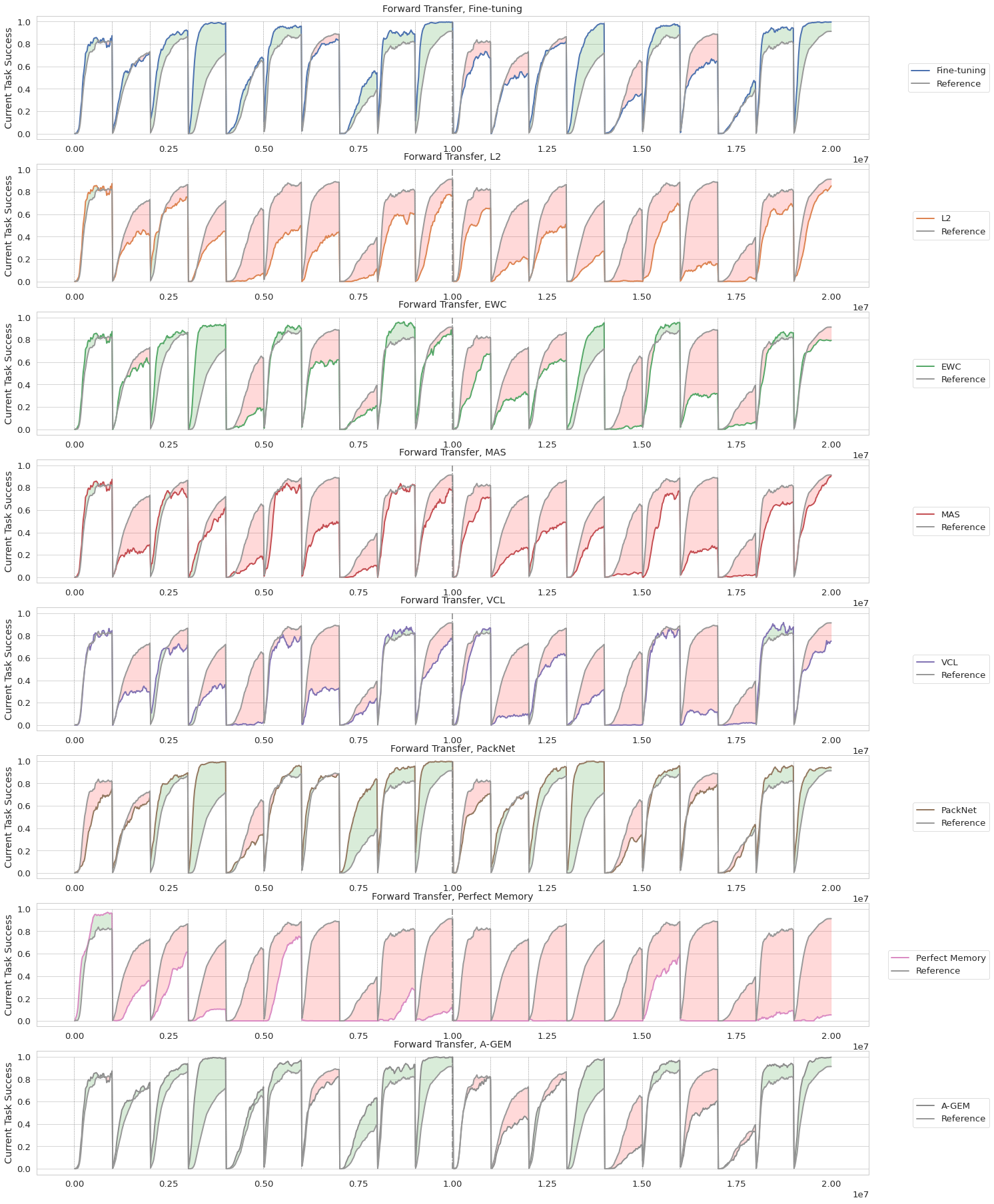}
        \caption{Forward transfer, the reference curves come from the single task training. }\label{fig:foraward_transfer_per_task}
    \end{center}
\end{figure}

Table \ref{tab:transfer_table} and Figure \ref{fig:transfer_heatmap} supplement Table \ref{tab:collected_resuls_appendix} with detailed results on forward transfer on \cwtwenty{}. Corresponding training curves can be find in Figure \ref{fig:foraward_transfer_per_task} and Figure \ref{fig:current_task_big}.

Fine-tuning is a strong baseline, with PackNet and AGEM taking close second and third place (although the latter method has a low overall performance). Unfortunately, all the other methods suffer from the negative forward transfer, which in all cases, except possibly from EWC, is quite significant. To make the picture grimmer, transfer on the second part of \cwtwenty{} is worse, even though the task has been learned (and not forgotten) on the first part, see Table \ref{tab:negative_transfer_delta}.

% \kucil{Average transfer over methods, for each task: [0.08, -0.26, 0.07, 0.21, -0.26, -0.43, -1.11, 0.0, -0.15, -0.26, -0.54, -0.57, -0.45, 0.02, -0.38, -0.46, -1.74, -0.14, -0.25, -0.26]}

\begin{table}[h]
    \centering
    \scriptsize 
    \rotatebox{90}{
    \begin{tabular}{llllllllll}
        \toprule
        Method & Fine-tuning                 & L2                           & EWC                          & MAS                          & VCL                          & PackNet                     & Perfect Memory               & A-GEM                     & mean  \\
        Task   &                             &                              &                              &                              &                              &                             &                              &                           &       \\
        \midrule
        0      & 0.13 [0.05, 0.20]           & 0.14 [0.07, 0.21]            & 0.13 [0.05, 0.20]            & 0.13 [0.05, 0.20]            & 0.02 [-0.09, 0.12]           & -0.49 [-0.65, -0.34]        & 0.43 [0.31, 0.54]            & 0.13 [0.04, 0.22]         & 0.08  \\
        1      & 0.06 [-0.09, 0.18]          & -0.39 [-0.53, -0.25]         & -0.10 [-0.19, -0.02]         & -0.63 [-0.72, -0.54]         & -0.51 [-0.60, -0.40]         & -0.08 [-0.19, 0.03]         & -0.68 [-0.83, -0.52]         & 0.21 [0.14, 0.28]         & -0.26 \\
        2      & 0.39 [0.23, 0.52]           & -0.09 [-0.40, 0.19]          & 0.35 [0.20, 0.49]            & 0.04 [-0.20, 0.24]           & -0.09 [-0.36, 0.14]          & 0.33 [0.09, 0.51]           & -0.82 [-1.18, -0.51]         & 0.48 [0.37, 0.58]         & 0.07  \\
        3      & 0.76 [0.73, 0.79]           & -0.26 [-0.53, -0.01]         & 0.66 [0.51, 0.76]            & -0.09 [-0.31, 0.13]          & -0.33 [-0.58, -0.12]         & 0.80 [0.76, 0.83]           & -0.71 [-0.91, -0.53]         & 0.81 [0.78, 0.83]         & 0.21  \\
        4      & 0.08 [-0.05, 0.20]          & -0.45 [-0.52, -0.38]         & -0.36 [-0.45, -0.26]         & -0.35 [-0.45, -0.25]         & -0.48 [-0.54, -0.42]         & -0.21 [-0.33, -0.09]        & -0.49 [-0.55, -0.43]         & 0.18 [0.05, 0.29]         & -0.26 \\
        5      & 0.55 [0.50, 0.59]           & -1.84 [-2.39, -1.29]         & 0.14 [0.04, 0.24]            & -0.44 [-0.59, -0.29]         & -0.38 [-0.58, -0.20]         & 0.05 [-0.22, 0.29]          & -1.96 [-2.41, -1.49]         & 0.45 [0.30, 0.57]         & -0.43 \\
        6      & -0.11 [-0.24, 0.00]         & -1.69 [-1.94, -1.44]         & -0.81 [-0.98, -0.62]         & -1.38 [-1.60, -1.17]         & -1.81 [-1.97, -1.65]         & 0.12 [0.04, 0.20]           & -2.92 [-3.05, -2.80]         & -0.27 [-0.38, -0.17]      & -1.11 \\
        7      & 0.13 [0.04, 0.22]           & -0.20 [-0.26, -0.15]         & -0.10 [-0.19, -0.01]         & -0.18 [-0.24, -0.12]         & -0.10 [-0.17, -0.03]         & 0.44 [0.34, 0.53]           & -0.24 [-0.29, -0.19]         & 0.28 [0.14, 0.42]         & 0.00  \\
        8      & 0.44 [0.38, 0.50]           & -0.91 [-1.45, -0.38]         & 0.35 [0.21, 0.47]            & -0.18 [-0.56, 0.14]          & 0.12 [-0.17, 0.35]           & 0.45 [0.29, 0.58]           & -1.86 [-2.20, -1.53]         & 0.41 [0.27, 0.54]         & -0.15 \\
        9      & 0.77 [0.70, 0.81]           & -0.68 [-1.20, -0.25]         & -0.11 [-0.64, 0.32]          & -0.51 [-1.05, -0.05]         & -0.66 [-1.21, -0.17]         & 0.80 [0.72, 0.86]           & -2.44 [-3.05, -1.95]         & 0.79 [0.73, 0.84]         & -0.26 \\
        10     & -0.30 [-0.60, -0.01]        & -0.49 [-0.82, -0.15]         & -0.66 [-0.96, -0.36]         & -0.41 [-0.74, -0.10]         & -0.07 [-0.35, 0.19]          & -0.41 [-0.74, -0.10]        & -1.73 [-1.85, -1.63]         & -0.23 [-0.67, 0.16]       & -0.54 \\
        11     & -0.21 [-0.30, -0.12]        & -0.77 [-0.90, -0.65]         & -0.57 [-0.70, -0.45]         & -0.71 [-0.81, -0.61]         & -0.91 [-1.01, -0.80]         & 0.08 [-0.02, 0.17]          & -1.07 [-1.15, -1.00]         & -0.43 [-0.61, -0.25]      & -0.57 \\
        12     & 0.04 [-0.14, 0.20]          & -0.74 [-1.09, -0.43]         & -0.36 [-0.68, -0.08]         & -0.67 [-1.00, -0.38]         & -0.38 [-0.66, -0.13]         & 0.42 [0.32, 0.50]           & -1.76 [-2.11, -1.48]         & -0.19 [-0.47, 0.06]       & -0.45 \\
        13     & 0.53 [0.44, 0.60]           & -0.59 [-0.79, -0.41]         & 0.39 [0.25, 0.51]            & -0.33 [-0.55, -0.13]         & -0.44 [-0.67, -0.22]         & 0.82 [0.79, 0.85]           & -0.78 [-0.96, -0.64]         & 0.56 [0.44, 0.67]         & 0.02  \\
        14     & -0.19 [-0.31, -0.05]        & -0.49 [-0.55, -0.43]         & -0.47 [-0.53, -0.41]         & -0.45 [-0.53, -0.37]         & -0.49 [-0.55, -0.43]         & -0.24 [-0.35, -0.13]        & -0.49 [-0.55, -0.43]         & -0.25 [-0.41, -0.09]      & -0.38 \\
        15     & 0.48 [0.41, 0.56]           & -1.52 [-1.99, -1.08]         & 0.03 [-0.13, 0.19]           & -1.05 [-1.40, -0.73]         & -0.20 [-0.41, -0.01]         & 0.17 [-0.00, 0.34]          & -1.94 [-2.44, -1.44]         & 0.36 [0.24, 0.48]         & -0.46 \\
        16     & -0.79 [-0.96, -0.63]        & -2.44 [-2.64, -2.26]         & -1.83 [-2.07, -1.59]         & -2.07 [-2.28, -1.85]         & -2.50 [-2.68, -2.33]         & -0.34 [-0.48, -0.19]        & -2.92 [-3.05, -2.80]         & -1.05 [-1.23, -0.86]      & -1.74 \\
        17     & 0.02 [-0.07, 0.11]          & -0.23 [-0.28, -0.18]         & -0.20 [-0.26, -0.13]         & -0.22 [-0.28, -0.17]         & -0.22 [-0.28, -0.17]         & -0.03 [-0.12, 0.06]         & -0.24 [-0.29, -0.19]         & -0.03 [-0.13, 0.09]       & -0.14 \\
        18     & 0.52 [0.45, 0.58]           & -0.74 [-1.23, -0.30]         & -0.06 [-0.46, 0.28]          & -0.67 [-1.19, -0.17]         & 0.26 [-0.04, 0.49]           & 0.44 [0.28, 0.58]           & -2.24 [-2.57, -1.86]         & 0.52 [0.46, 0.58]         & -0.25 \\
        19     & 0.73 [0.66, 0.78]           & -0.59 [-1.08, -0.18]         & -0.16 [-0.71, 0.29]          & -0.26 [-0.77, 0.13]          & -0.45 [-1.02, 0.03]          & 0.46 [0.12, 0.70]           & -2.49 [-3.11, -1.98]         & 0.68 [0.58, 0.75]         & -0.26 \\
        \midrule
        mean   & {0.20 {\tiny [0.17, 0.23]}} & -0.75 {\tiny [-0.87, -0.65]} & -0.19 {\tiny [-0.25, -0.14]} & -0.52 {\tiny [-0.58, -0.48]} & -0.48 {\tiny [-0.56, -0.42]} & {0.18 {\tiny [0.14, 0.21]}} & -1.37 {\tiny [-1.46, -1.30]} & 0.17 {\tiny [0.13, 0.20]} & ---   \\
        \bottomrule
    \end{tabular}}
    \caption{Forward transfers on \cwtwenty{} (rows) for each method (columns).} \label{tab:transfer_table}
    %\label{tab:my_label}
\end{table}

\begin{table}[h]
    \centering\small
    \begin{tabular}{lrrrrrrrrrr}
        \toprule
        Task           & $\Delta_0$ & $\Delta_1$ & $\Delta_2$ & $\Delta_3$ & $\Delta_4$ & $\Delta_5$ & $\Delta_6$ & $\Delta_7$ & $\Delta_8$ & $\Delta_9$ \\
        Method         &            &            &            &            &            &            &            &            &            &            \\
        \midrule
        Fine-tuning    & -0.42      & -0.27      & -0.34      & -0.24      & -0.26      & -0.06      & -0.68      & -0.11      & 0.08       & -0.04      \\
        L2             & -0.63      & -0.39      & -0.66      & -0.33      & -0.03      & 0.32       & -0.76      & -0.02      & 0.17       & 0.10       \\
        EWC            & -0.79      & -0.47      & -0.71      & -0.27      & -0.11      & -0.11      & -1.02      & -0.09      & -0.41      & -0.05      \\
        MAS            & -0.54      & -0.08      & -0.71      & -0.24      & -0.10      & -0.61      & -0.68      & -0.04      & -0.49      & 0.25       \\
        VCL            & -0.08      & -0.40      & -0.29      & -0.11      & -0.01      & 0.18       & -0.69      & -0.13      & 0.13       & 0.20       \\
        PackNet        & 0.09       & 0.16       & 0.08       & 0.02       & -0.03      & 0.12       & -0.46      & -0.47      & -0.01      & -0.34      \\
        Perfect Memory & -2.16      & -0.39      & -0.94      & -0.07      & 0.00       & 0.02       & 0.00       & 0.00       & -0.38      & -0.05      \\
        A-GEM          & -0.36      & -0.63      & -0.67      & -0.25      & -0.43      & -0.09      & -0.77      & -0.32      & 0.11       & -0.11      \\
        \bottomrule
    \end{tabular}
    \caption{Difference in transfer when revisiting tasks. $\Delta_i = \text{FT}_{i+10} - \text{FT}_i$; recall that $i$ and $i+10$ are the same tasks in \cwtwenty{}.} \label{tab:negative_transfer_delta}
    %\label{tab:my_label}
\end{table}

\subsection{Results for CW10}
In Table~\ref{tab:cw10} we present results for the CW10 sequence ($10$ task version of the benchmark, without repeating the tasks). One can see that results are mostly consistent with \cwtwenty{} (see Table \ref{tab:collected_resuls_appendix}), and thus CW10 can be used for faster experimenting. More detailed transfer results can be found in Table \ref{tab:transfer_table}.

\begin{table}[th]
    \begin{center}
        \tabcolsep=0.11cm
        \renewcommand{\arraystretch}{0.9}
        \begin{tabular}{llll}
\toprule
 method   & performance       & forgetting           & f. transfer   \\
\midrule
 \textbf{Fine-tuning}        & 0.10 {\tiny [0.10, 0.11]} & 0.74 {\tiny [0.72, 0.75]}    & \textbf{0.32 {\tiny [0.28, 0.35]}}     \\
 \textbf{L2}          & 0.48 {\tiny [0.43, 0.53]} & 0.02 {\tiny [-0.00, 0.04]}   & -0.57 {\tiny [-0.77, -0.39]}  \\
 \textbf{EWC}         & 0.66 {\tiny [0.62, 0.69]} & 0.03 {\tiny [0.01, 0.06]}    & 0.05 {\tiny [-0.02, 0.12]}    \\
 \textbf{MAS}         & 0.59 {\tiny [0.56, 0.61]} & -0.02 {\tiny [-0.03, -0.00]} & -0.35 {\tiny [-0.42, -0.28]}  \\
 \textbf{VCL}         & 0.53 {\tiny [0.49, 0.58]} & -0.02 {\tiny [-0.03, -0.00]} & -0.44 {\tiny [-0.57, -0.32]}  \\
 \textbf{PackNet}     & \textbf{0.83 {\tiny [0.81, 0.85]}} & -0.00 {\tiny [-0.01, 0.01]}  & 0.21 {\tiny [0.16, 0.25]}     \\
 \textbf{Perfect Memory}   & 0.29 {\tiny [0.26, 0.31]} & 0.03 {\tiny [0.02, 0.05]}    & -1.11 {\tiny [-1.20, -1.04]}  \\
 \textbf{A-GEM}        & 0.13 {\tiny [0.12, 0.14]} & 0.73 {\tiny [0.72, 0.75]}    & \textbf{0.29 {\tiny [0.27, 0.32]}}     \\
 \midrule
 \textbf{MT}         & 0.51 {\tiny [0.48, 0.53]} & ---                  & ---                   \\
 \textbf{MT (PopArt)}  & 0.65 {\tiny [0.63, 0.67]} & ---                  & ---                   \\
            \bottomrule
        \end{tabular}
        \caption{Results on CW10, for CL methods and multi-task training.}
        \label{tab:cw10}
    \end{center}
\end{table}

\clearpage
\newpage

\section{Triplet experiments}\label{sec:triplets_appendix}
In addition to the CW10 and CW20 sequences, we propose eight triplets. They are sequences of three tasks aimed to allow rapid experimenting. The triplets were selected in order to capture situations when remembering is crucial for forward transfer. That is, we consider tasks $A \rightarrow B \rightarrow C$, where $A \rightarrow C$ exhibits bigger forward transfer than $B \rightarrow C$ (based on the transfer matrix), see Table \ref{tab:triplet_transfers} . We propose the following triplets:

\begin{enumerate}
    \item  $\texttt{push-v1}\rightarrow \texttt{window-close-v1} \rightarrow \texttt{hammer-v1}$
    \item  $\texttt{hammer-v1} \rightarrow \texttt{window-close-v1} \rightarrow \texttt{faucet-close-v1} $
    \item  $\texttt{stick-pull-v1} \rightarrow \texttt{push-back-v1} \rightarrow \texttt{push-wall-v1} $
    \item  $\texttt{push-wall-v1} \rightarrow \texttt{shelf-place-v1} \rightarrow \texttt{push-back-v1}$
    \item  $\texttt{faucet-close-v1} \rightarrow \texttt{shelf-place-v1} \rightarrow \texttt{push-back-v1}$
    \item  $\texttt{stick-pull-v1} \rightarrow \texttt{peg-unplug-side-v1} \rightarrow \texttt{stick-pull-v1}$
    \item  $\texttt{window-close-v1} \rightarrow \texttt{handle-press-side-v1} \rightarrow \texttt{peg-unplug-side-v1}$
    \item  $\texttt{faucet-close-v1} \rightarrow \texttt{shelf-place-v1} \rightarrow \texttt{peg-unplug-side-v1}$
\end{enumerate}

\begin{figure}
    \centering
    \includegraphics[width=\textwidth]{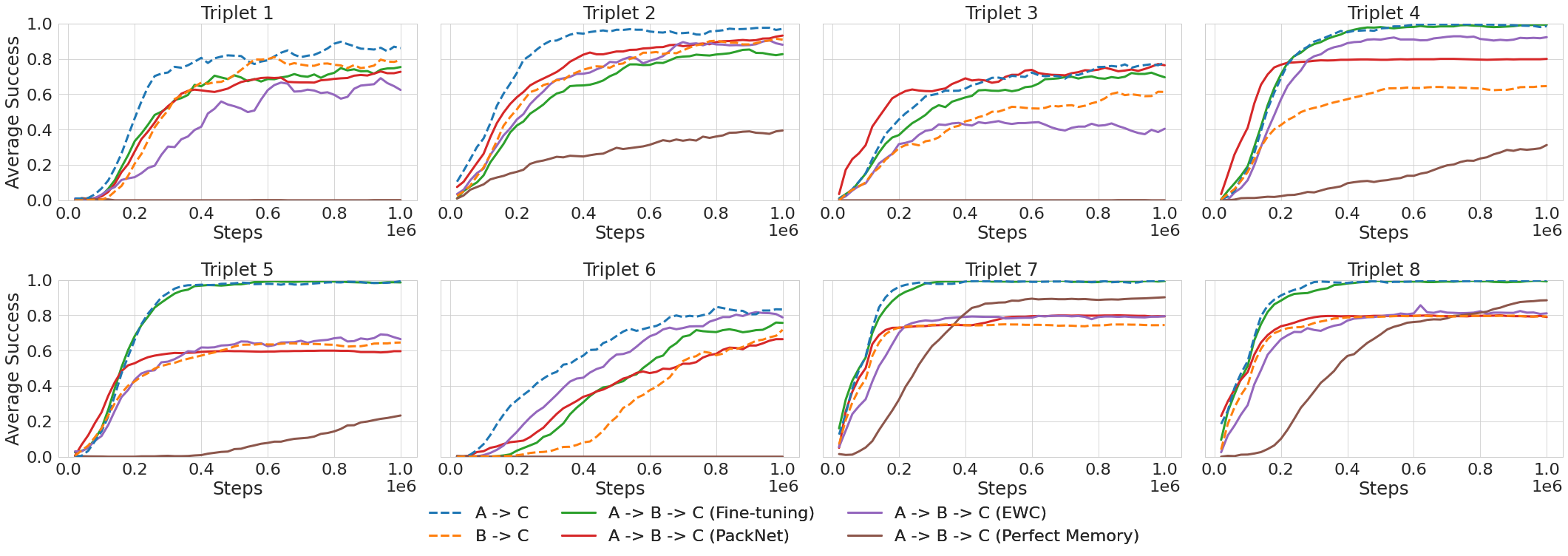}
    \caption{Performance of selected methods on the triplets sequences.}
    \label{fig:more_triples_appendix}
\end{figure}

In Figure \ref{fig:more_triples_appendix} and Table \ref{tab:triplet_transfers} we present evaluations for the third task. On most of the triplets, the tested methods  (Fine-tuning, EWC, PackNet, Perfect Memory) are not able to reach the performance obtained by direct transfer $A \rightarrow C$. Moreover, Fine-tuning often outperforms the rest of the methods. Interestingly, in triplets 7 and 8, Fine-tuning is able to achieve transfer comparable to $A \rightarrow C$, while EWC and PackNet smaller one similar to $B \rightarrow C$. Perfect memory does not perform well on any triplet, which is consistent with our experiments on the long sequences.

\begin{table}
    \centering\scriptsize
    \begin{tabular}{lp{1cm}p{1cm}p{2cm}p{2cm}p{2cm}p{2.3cm}}
        \toprule
        Triplet & A$\rightarrow$ C & B $\rightarrow$ C & A $\rightarrow$ B $\rightarrow$ C (Fine-tuning) & A $\rightarrow$ B $\rightarrow$ C (PackNet) & A $\rightarrow$ B $\rightarrow$ C (EWC) & A $\rightarrow$ B $\rightarrow$ C (Perfect Mem.) \\
        \midrule
        1       & 0.20             & -0.10             & -0.14                                           & -0.21                                       & -0.50                                   & -1.73                                            \\
        2       & 0.64             & 0.24              & 0.06                                            & 0.36                                        & 0.20                                    & -0.98                                            \\
        3       & 0.21             & -0.15             & 0.10                                            & 0.28                                        & -0.31                                   & -1.09                                            \\
        4       & 0.76             & 0.20              & 0.76                                            & 0.56                                        & 0.62                                    & -0.52                                            \\
        5       & 0.75             & 0.20              & 0.76                                            & 0.20                                        & 0.22                                    & -0.64                                            \\
        6       & 0.42             & -0.02             & 0.14                                            & 0.10                                        & 0.29                                    & -0.49                                            \\
        7       & 0.80             & -0.01             & 0.79                                            & 0.11                                        & 0.06                                    & -0.00                                            \\
        8       & 0.79             & 0.10              & 0.74                                            & 0.15                                        & 0.05                                    & -0.46                                            \\
        \midrule
        mean    & 0.57             & 0.06              & 0.40                                            & 0.19                                        & 0.08                                    & -0.74                                            \\
        \bottomrule
    \end{tabular}
    \caption{Forward Transfers for triplets.}\label{tab:triplet_transfers}
\end{table}

\section{Ablations} \label{sec:ablations_appendix}
\subsection{Other orders of tasks}
\label{app:abl_order}
\begin{figure}[h!]
    \begin{center}
        \includegraphics[width=\textwidth]{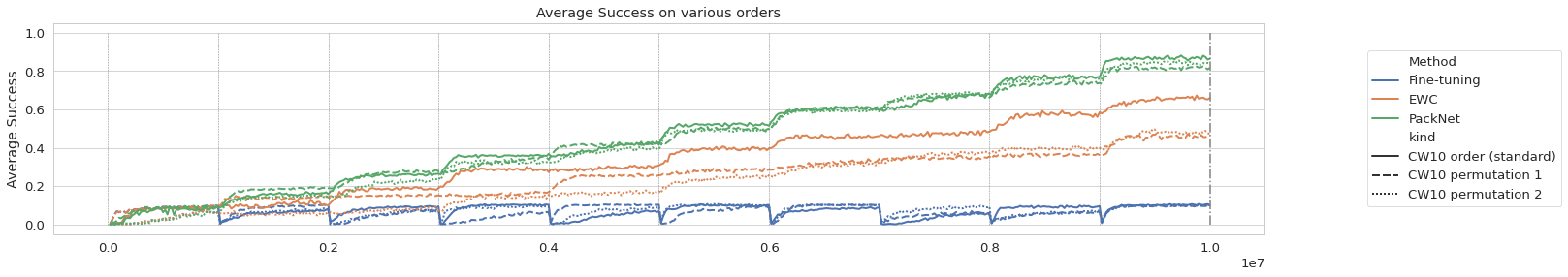}
        \caption{\label{fig:abl_order} Average performance for selected methods on CW10 and its two random orderings.}
    \end{center}
\end{figure}

We study the impact of task order on the performance of CL methods. In Figure~\ref{fig:abl_order}, we show average performance curves for three selected CL methods that are run with the same hyperparameters on three different orderings: the  CW10 sequence (see Section \ref{sec:sequences_appendix}) and its two random permutations.
Further, Table~\ref{tab:ablation_order} contains values of the forgetting and forward transfer metrics.

The ranking of the methods is preserved; we also observe that forward transfer of EWC varies substantially with the change of sequence order. To a lesser extent, this also impacts the performance metric. Decreasing this variance is, in our view, an important research question. We speculate that it is much related to improving transfer in general. 

\begin{table}[ht!]
    \begin{center}
        \tabcolsep=0.11cm
        \renewcommand{\arraystretch}{0.9}
        \begin{tabular}{llll}
            \toprule
            method                                      & performance               & forgetting                 & f. transfer                  \\
            \midrule
            \textbf{Fine-tuning, CW10 order (standard)} & 0.10 {\tiny [0.10, 0.11]} & 0.74 {\tiny [0.72, 0.75]}  & 0.32 {\tiny [0.29, 0.35]}    \\
            \textbf{EWC, CW10 order (standard)}         & 0.66 {\tiny [0.63, 0.69]} & 0.03 {\tiny [0.01, 0.06]}  & 0.02 {\tiny [-0.05, 0.07]}   \\
            \textbf{PackNet, CW10 order (standard)}     & 0.87 {\tiny [0.85, 0.89]} & 0.01 {\tiny [-0.01, 0.02]} & 0.38 {\tiny [0.36, 0.41]}    \\
            \midrule
            \textbf{Fine-tuning, CW10 permutation 1}    & 0.10 {\tiny [0.10, 0.10]} & 0.72 {\tiny [0.69, 0.74]}  & 0.18 {\tiny [0.13, 0.23]}    \\
            \textbf{EWC, CW10 permutation 1}            & 0.46 {\tiny [0.43, 0.50]} & 0.02 {\tiny [0.00, 0.04]}  & -0.45 {\tiny [-0.50, -0.41]} \\
            \textbf{PackNet, CW10 permutation 1}        & 0.82 {\tiny [0.80, 0.84]} & 0.03 {\tiny [0.01, 0.05]}  & 0.25 {\tiny [0.19, 0.30]}    \\
            \midrule
            \textbf{Fine-tuning, CW10 permutation 2}    & 0.10 {\tiny [0.10, 0.10]} & 0.73 {\tiny [0.72, 0.75]}  & 0.22 {\tiny [0.19, 0.26]}    \\
            \textbf{EWC, CW10 permutation 2}            & 0.48 {\tiny [0.43, 0.52]} & 0.05 {\tiny [0.03, 0.07]}  & -0.54 {\tiny [-0.68, -0.42]} \\
            \textbf{PackNet, CW10 permutation 2}        & 0.84 {\tiny [0.81, 0.86]} & 0.01 {\tiny [0.00, 0.03]}  & 0.30 {\tiny [0.26, 0.34]}    \\
            \bottomrule
        \end{tabular}
        \caption{Results on the CW10 sequence and its 2 random permutations.}
        \label{tab:ablation_order}
    \end{center}
\end{table}

The new sequences are:

{\small
\texttt{handle-press-side-v1, faucet-close-v1, shelf-place-v1,  stick-pull-v1, peg-unplug-side-v1, hammer-v1, push-back-v1, push-wall-v1, push-v1, window-close-v1}}

and

    {\small \texttt{stick-pull-v1, push-wall-v1, shelf-place-v1,
            window-close-v1, hammer-v1, peg-unplug-side-v1,
            push-back-v1, faucet-close-v1, push-v1,
            handle-press-side-v1.}}

\subsection{Random order of tasks}\label{subsec:random_orders}

We further study the impact of task orderings by evaluating on $20$ different random orders, see Table~\ref{tab:random_orders}. Obtained results are qualitatively similar to the ones with the fixed ordering.

%We generate 20 random permutations of CW10.
% In  we report results for all the methods. Each metric is computed as an average over 20 different random orders of CW10.

\begin{table}[ht!]
    \begin{center}
        \tabcolsep=0.11cm
        \renewcommand{\arraystretch}{0.9}
        \begin{tabular}{llll}
\toprule
 method   & performance       & forgetting          & f. transfer   \\
\midrule
 \textbf{Fine-tuning}        & 0.08 {\tiny [0.07, 0.09]} & 0.71 {\tiny [0.68, 0.74]}   & 0.17 {\tiny [0.09, 0.24]}     \\
 \textbf{L2}          & 0.42 {\tiny [0.34, 0.49]} & 0.00 {\tiny [-0.01, 0.01]}  & -0.73 {\tiny [-0.90, -0.56]}  \\
 \textbf{EWC}         & 0.53 {\tiny [0.48, 0.58]} & 0.01 {\tiny [-0.01, 0.03]}  & -0.31 {\tiny [-0.43, -0.18]}  \\
 \textbf{MAS}         & 0.41 {\tiny [0.35, 0.47]} & 0.00 {\tiny [-0.02, 0.02]}  & -0.65 {\tiny [-0.79, -0.51]}  \\
 \textbf{VCL}         & 0.39 {\tiny [0.34, 0.44]} & 0.02 {\tiny [-0.00, 0.03]}  & -0.69 {\tiny [-0.80, -0.58]}  \\
 \textbf{PackNet}     & 0.72 {\tiny [0.68, 0.76]} & 0.00 {\tiny [-0.01, 0.01]}  & -0.04 {\tiny [-0.14, 0.05]}   \\
 \textbf{Perfect Memory}   & 0.27 {\tiny [0.24, 0.31]} & -0.00 {\tiny [-0.02, 0.01]} & -1.12 {\tiny [-1.22, -1.02]}  \\
 \textbf{A-GEM}        & 0.10 {\tiny [0.08, 0.11]} & 0.70 {\tiny [0.67, 0.73]}   & 0.19 {\tiny [0.12, 0.24]}     \\
\bottomrule
\end{tabular}
        \caption{Results averaged over 20 different random orderings of CW10 -- see Subsection~\ref{subsec:random_orders}}
        \label{tab:random_orders}
    \end{center}
\end{table}

\subsection{One-hot inputs}
\begin{figure}
    \begin{center}
        \includegraphics[width=\textwidth]{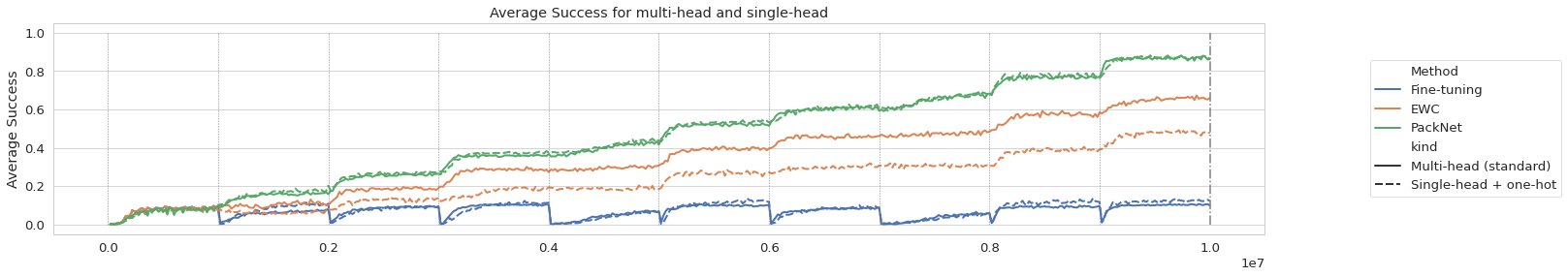}
        \caption{\label{fig:abl_sh} Average performance for selected CL methods on CW10 with standard multi-head architecture and single-head architecture with one-hot task encoding added to input.}
    \end{center}
\end{figure}

We study an alternative setup in which a single-head architecture is used, and one-hot encoding is provided in input for the task identification. In Table \ref{tab:ablation_sh} and Figure~\ref{fig:abl_sh} we present results for three selected CL methods in both the multi-head and single-head setups. We see that the performance of EWC drops a bit, while PackNet remains mostly unaffected. Table~\ref{tab:ablation_sh} contains values of average performance, forgetting, and forward transfer computed in this setting.

For the single-head experiments, we ran another hyperparameter search. For the single-head EWC experiment the selected value is $\lambda = 10^3$ as opposed to $\lambda = 10^4$ for multi-head. We conjecture that the multi-head case can be regularized more aggressively, as for each task, it gets a small number of new parameters (in new heads).

\begin{table}[ht!]
    \begin{center}
        \tabcolsep=0.11cm
        \renewcommand{\arraystretch}{0.9}
        \begin{tabular}{llll}
            \begin{tabular}{llll}
                \toprule
                method                                       & performance               & forgetting                  & f. transfer                  \\
                \midrule
                \textbf{Fine-tuning, Multi-head (standard) } & 0.10 {\tiny [0.10, 0.11]} & 0.74 {\tiny [0.72, 0.75]}   & 0.32 {\tiny [0.29, 0.35]}    \\
                \textbf{EWC, Multi-head (standard)}          & 0.66 {\tiny [0.63, 0.69]} & 0.03 {\tiny [0.01, 0.06]}   & 0.02 {\tiny [-0.05, 0.07]}   \\
                \textbf{PackNet, Multi-head (standard)}      & 0.87 {\tiny [0.85, 0.89]} & 0.01 {\tiny [-0.01, 0.02]}  & 0.38 {\tiny [0.36, 0.41]}    \\
                \midrule
                \textbf{Fine-tuning, Single-head + one-hot } & 0.12 {\tiny [0.12, 0.13]} & 0.68 {\tiny [0.66, 0.70]}   & 0.16 {\tiny [0.12, 0.19]}    \\
                \textbf{EWC, Single-head + one-hot}          & 0.48 {\tiny [0.45, 0.50]} & 0.14 {\tiny [0.12, 0.15]}   & -0.25 {\tiny [-0.33, -0.17]} \\
                \textbf{PackNet, Single-head + one-hot}      & 0.87 {\tiny [0.85, 0.89]} & -0.01 {\tiny [-0.02, 0.01]} & 0.24 {\tiny [0.20, 0.28]}    \\
                \bottomrule
            \end{tabular}
        \end{tabular}
        \caption{Results on CW10 with original multi-head setup and alternative with single-head and one-hot encoding}
        \label{tab:ablation_sh}
    \end{center}
\end{table}

\label{sec:onehot}

\subsection{Experiments with a sequence of $30$ tasks}

An important question for continual learning is how methods scale to a larger number of tasks. To test this in our setting, we consider a sequence of $30$ tasks. To select challenging but solvable tasks we removed $10$ easiest and $10$ hardest task of MetaWorld (measured by performance in single-task learning).

% An intriguing question in continual learning is how methods scale to a larger number of tasks. To test this in our setting, we consider a sequence of $30$ tasks. The sequence is obtained by taking the base $50$ tasks from MetaWorld and removing $10$ easiest and $10$ hardest tasks, as measured by performance of a single-task learner. This is done in order to avoid training on tasks which are either nearly impossible or trivial to learn, which allows us to focus on tasks which are challenging, but still solvable.

This way we obtained a set of $30$ following tasks:
{\small \textsc{plate-slide-v1}, \textsc{plate-slide-back-side-v1}, \textsc{handle-press-v1}, \textsc{handle-pull-v1}, \textsc{handle-pull-side-v1}, \textsc{soccer-v1}, \textsc{coffee-push-v1}, \textsc{coffee-button-v1}, \textsc{sweep-into-v1}, \textsc{dial-turn-v1}, \textsc{hand-insert-v1}, \textsc{window-open-v1}, \textsc{plate-slide-side-v1}, \textsc{plate-slide-back-v1}, \textsc{door-lock-v1}, \textsc{door-unlock-v1}, \textsc{push-v1}, \textsc{door-open-v1}, \textsc{box-close-v1}, \textsc{faucet-open-v1}, \textsc{coffee-pull-v1}, \textsc{shelf-place-v1}, \textsc{faucet-close-v1}, \textsc{handle-press-side-v1}, \textsc{push-wall-v1}, \textsc{sweep-v1}, \textsc{stick-push-v1}, \textsc{bin-picking-v1}, \textsc{basketball-v1}, \textsc{hammer-v1}}.

We then train four methods: fine-tuning, EWC, PackNet, and Perfect Memory on this set of tasks. We use the random ordering approach presented in Subsection \ref{subsec:random_orders} with 40 seeds to obtain reliable results. The tested methods perform similarly as in the case of CW20, see Table \ref{tab:ablation_cw30}.  PackNet significantly outperforms the rest of the methods, followed by EWC. Reservoir and fine-tuning are affected with catastrophic forgetting, with performance showing that they are mostly able to solve a single task. %Since the base setup of $20$ tasks shows similar findings with reduced computational complexity, we decide to use this setting as our main one.

\begin{table}[ht!]
    \begin{center}
        \begin{tabular}{crrrr}
            \toprule
               & Fine-tuning   & EWC & PackNet & Perfect Memory  \\
            \midrule%\small 
            \textbf{Performance} & 0.04 {\small [0.03, 0.04]} & 0.44 {\small [0.42, 0.46]}   & 0.70 {\small [0.68, 0.71]} & 0.04 {\small [0.03, 0.04]} \\
            \bottomrule
        \end{tabular}
        \caption{Results on 30 tasks.}
        \label{tab:ablation_cw30}
    \end{center}
\end{table}

\section{Multi-task learning}\label{sec:mult_task_learning_appendix}
\begin{figure}
    \begin{center}
        \includegraphics[width=\textwidth]{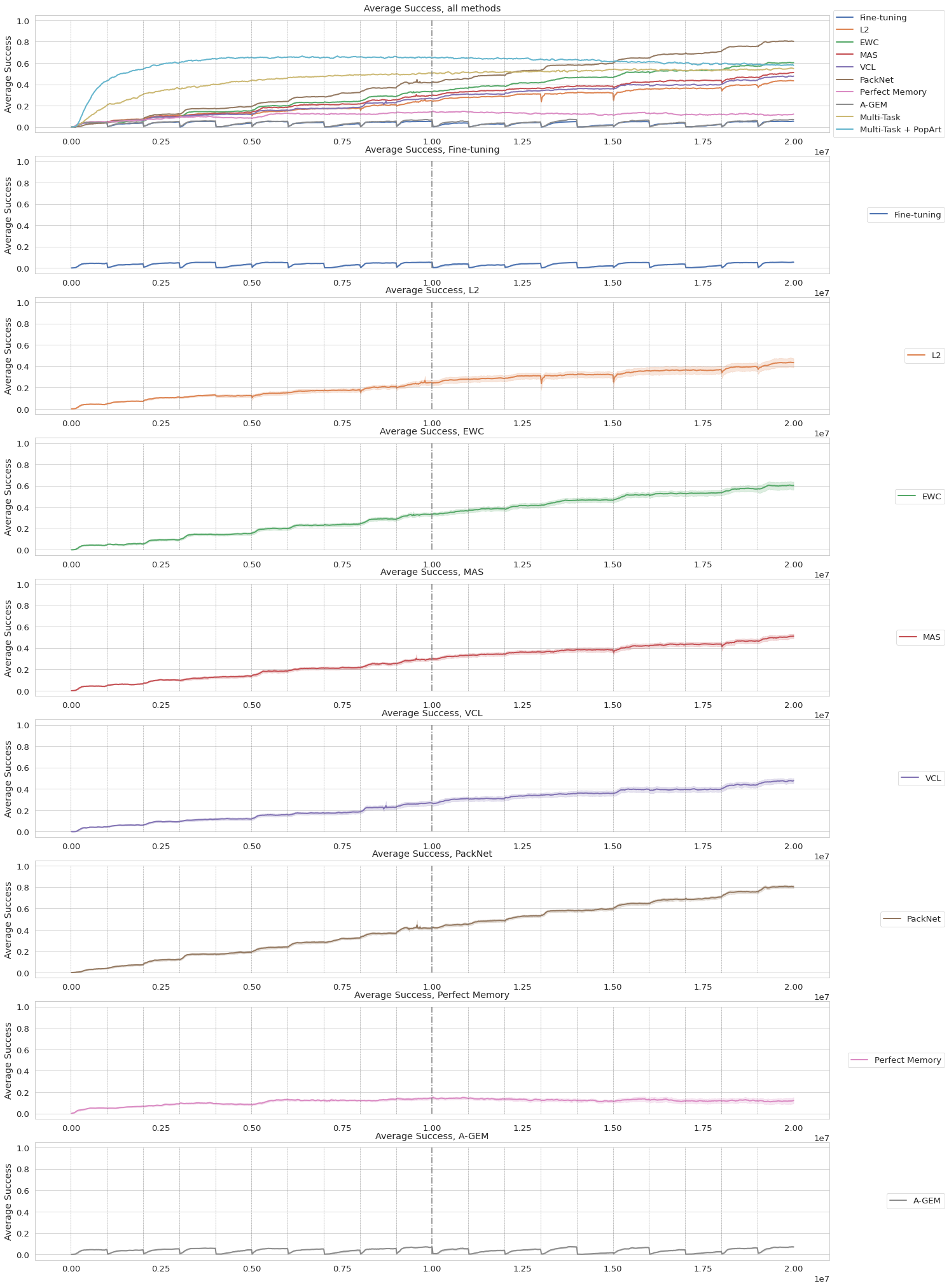}
        \caption{Average (over tasks) success rate for all tested methods and multi-task training.}\label{sec:average_success_big}
    \end{center}
\end{figure}

In multi-task experiments, we train $10$ tasks from CW10 simultaneously (we do not duplicate tasks as it is done in CW20). In our experiments, PopArt reward normalization~\cite{DBLP:conf/aaai/HesselSE0SH19} yields significant improvements ($0.66$ vs $0.50$ for the standard version, see also the top graph in  Figure~\ref{sec:average_success_big}). We observe that EWC and PackNet achieve results similar or better to multi-task learning. We consider this to be an interesting research direction. We conjecture that CL methods might benefit from the fact that the critic network stores only one value function (corresponding to the current task), see also discussion in Section \ref{sec:critic_regularizaiton_appendix}.

For multi-task training, we tested the following hyperparameter values: $\text{batch size} \in \{ 128, 256, 512 \}$ (selected value $=128$), $\text{learning rate} \in \{ \num{3e-5}, \num{1e-4}, \num{3e-4}, \num{1e-3} \}$ (selected value $=\num{1e-4}$).

\section{Infrastructure used}
We ran our experiments on clusters with servers typically equipped with $24$ or $28$ CPU cores and $64$GB of memory (no GPU). A typical experiment CW20 experiment was $100$ hours long and used $8$ or $12$ CPU cores.

During the project, we run more than $45$K experiments.

\newpage

% \section*{Further todos (after the deadline)}
% Webpage, possibly establish slack, or join the meta-world slack. \razp{Clean/prepare code for release.}

% TODO
% \section{Time and Memory}

\end{document}